\documentclass[preprint,10pt]{elsarticle}

\usepackage{epstopdf}
\usepackage{amssymb}

\usepackage{amsmath}
\usepackage{amsfonts}
\usepackage{amssymb}
\usepackage{graphicx}
\usepackage{subfigure}
\usepackage{url}
\usepackage{tikz}
\usepackage{amsthm}
\usepackage{algorithm}
\usepackage{algorithmic}

\theoremstyle{definition}

\newcommand{\argmin}{\mathop{\arg \min}}

\newcommand{\dist}{\mathit{\Delta}}

\graphicspath{{./fig/}}

\journal{Physica A}

\begin{document}

\begin{frontmatter}



\title{Developments in the theory of randomized shortest paths with a comparison of graph node distances}


\author{Ilkka Kivim\"aki}
\address{ICTEAM institute, Universit\'e catholique de Louvain, Louvain-la-Neuve, Belgium}
\author{Masashi Shimbo}
\address{Nara Institute of Science and Technology, Ikoma, Japan}
\author{Marco Saerens}
\address{ICTEAM institute, Universit\'e catholique de Louvain, Louvain-la-Neuve, Belgium}

\begin{abstract}
There have lately been several suggestions for parametrized distances on a graph that generalize the shortest path distance and the commute time or resistance distance. The need for developing such distances has risen from the observation that the above-mentioned common distances in many situations fail to take into account the global structure of the graph. In this article, we develop the theory of one family of graph node distances, known as the randomized shortest path dissimilarity, which has its foundation in statistical physics. We show that the randomized shortest path dissimilarity can be easily computed in closed form for all pairs of nodes of a graph. Moreover, we come up with a new definition of a distance measure that we call the free energy distance. The free energy distance can be seen as an upgrade of the randomized shortest path dissimilarity as it defines a metric, in addition to which it satisfies the graph-geodetic property. The derivation and computation of the free energy distance are also straightforward. We then make a comparison between a set of generalized distances that interpolate between the shortest path distance and the commute time, or resistance distance. This comparison focuses on the applicability of the distances in graph node clustering and classification. The comparison, in general, shows that the parametrized distances perform well in the tasks. In particular, we see that the results obtained with the free energy distance are among the best in all the experiments.
\end{abstract}


\end{frontmatter}



\section{Introduction}
\label{sec:intro}
Defining distances and similarities between nodes of a graph based on its structure has become an essential task in the analysis of network data~\cite{Wasserman-1994, Thelwall04, chung06, Liben-Nowell-2007, kolaczyk09, Lewis09, newman10, Lu-2011, Estrada-2012}. In the simplest case, a binary network can be presented as an adjacency matrix or adjacency list which can be difficult to interpret. Acquiring meaningful information from such data requires sophisticated methods which often need to be chosen based on the context. Being able to measure the distance between the nodes of a network in a meaningful way of course provides a fundamental way of interpreting the network. With the information of distances between the nodes, one can apply traditional multivariate statistical or machine learning methods for analyzing the data.


The most common ways of defining a distance on a graph are to consider either the lengths of the shortest paths between nodes, leading to the definition of the \emph{shortest path (SP) distance}, or the expected lengths of random walks on the graph, which can be used to derive the \emph{commute time (CT) distance}~\cite{Gobel-1974}. The CT distance is known to equal the \emph{resistance distance}~\cite{Klein-1993, Snell-1984} up to a constant factor~\cite{Chandra-1989}. In this paper, we examine generalized distances on graphs that interpolate, depending on a parameter, between the shortest path distance and the commute time or resistance distance.

The paper contains several separate contributions: First, we develop the theory of one generalized distance, the \emph{randomized shortest path (RSP) dissimilarity}~\cite{RSP, Saerens-2008}. We derive a new algorithm for computing it for all pairs of nodes of a graph in closed form, and thus much more efficiently than before. We then derive another generalized distance from the RSP framework based on the Helmholtz free energy between two states of a thermodynamic system. We show that this \emph{free energy (FE) distance} actually coincides with the \emph{potential distance}, proposed in recent literature in a more ad hoc manner~\cite{BoP}. However, our new derivation gives a nice theoretical background for this distance. Finally, we make a comparison of the behavior and performance of different generalized graph node distances. The comparisons are conducted by observing the relative differences of distances between nodes in small example graphs and by examining the performance of the different distance measures in clustering and classification tasks.

The paper is structured as follows: In Section~\ref{sec:Terminology}, we define the terms and notation used in the paper. In our framework, we consider graphs where the edges are assigned weights and costs, which can be independent of each other. In Section~\ref{sec:DistFams}, we recall the definitions of the common distances on graphs. We also present a surprising result related to the generalization of the commute time distance considering costs, namely that the distance based on costs equals the commute time distance, up to a constant factor. In Section~\ref{sec:RSP}, we revisit the definition of the RSP dissimilarity~\cite{RSP, Saerens-2008}. We then derive the closed form algorithm, mentioned above, for computing it, and then formulate the definition of the FE distance. In Section~\ref{sec:OtherDists}, we present other parametrized distances on graphs interpolating between the SP and CT distances that have been defined in recent literature. Section~\ref{sec:Exp} contains the comparison of the RSP dissimilarity, the FE distance and the generalized distances defined in Section~\ref{sec:OtherDists}. Finally, Section~\ref{sec:Concl} sums up the content of the article.


\section{Terminology and notation}
\label{sec:Terminology}
We first go through the terminology and notation used in this paper. We denote by $G = (V,E)$ a graph $G$ consisting of a node set $V = \{1,2, \ldots , n\}$ and an edge set $E = \{ (i,j) \}$. Nodes $i$ and $j$ such that $(i,j)\in E$ are called \emph{adjacent} or \emph{connected}. Each graph can be represented as an adjacency matrix $\mathbf{A}$, where the elements $a_{ij}$ are called \emph{affinities}, or \emph{weights}, interchangeably. For unweighted graphs $a_{ij} = 1$ if $(i,j) \in E$, for weighted graphs $a_{ij} > 0$ if $(i,j) \in E$ and in both cases $a_{ij} = 0$ if $(i,j) \notin E$. The affinities can be interpreted as representing the degree of similarity between connected nodes. A \emph{path}, or \emph{walk}, interchangeably, on the graph $G$ is a sequence of nodes $\wp = (i_{0}, \ldots, i_{T})$, where $T \geq 0$ and $(i_{\tau},i_{\tau+1}) \in E$ for all $\tau = 0,\ldots,T-1$. The \emph{length} of the path, or walk, $\wp$, is then $T$. Note that throughout this article we include zero-length paths $(i), i \in V$ in the definition of a path, although in some contexts it may be more appropriate to disallow this by setting $T \geq 1$ in the definition. Moreover, we define \emph{absorbing}, or \emph{hitting} paths as paths which contain the terminal node only once. Thus a path $\wp$ is an absorbing path if $\wp = (i_{0}, \ldots i_{T})$, where $i_{T} \neq i_{\tau}$ for all $\tau = 0,\ldots,T-1$.

In addition to affinities, the edges of a graph can be assigned \emph{costs}, $c_{ij}$, such that $0 < c_{ij} < \infty$ if $(i,j)\in E$. The cost of a path $\wp$ is the sum of the costs along the path\footnote{Throughout the article we will use the tilde ($\sim$) to differentiate quantities related to paths from quantities related to edges.} $\widetilde{c}(\wp) = \sum_{(i,j) \in \wp}c_{ij}$. In principle, we do not define costs for unconnected pairs of nodes, but when making matrix computations, we assign the corresponding matrix elements a very large number (compared to other costs). When there is no natural cost assigned to the edges, a common convention is to define the costs as reciprocals of the affinities $c_{ij} = 1/a_{ij}$. This applies both for unweighted and weighted graphs. This way the edge weights and costs are analogous to conductance and resistance, respectively, in an electric network. In the experiments, in Section~\ref{sec:Exp} of this paper, we always use this conversion for determining costs from affinities. However, in the theory that we present in Sections~\ref{sec:DistFams}-\ref{sec:RSP}, we consider that the costs can also be assigned independently of the affinities, allowing a more general setting. This can be useful in many applications because links can often have a two-sided nature, on one hand based on the structure of the graph and on the other hand based on internal features of the edges. One such example can be a toll road network, where the affinities represent the proximities of places and the costs represent toll costs of traversing a road. This interpretation is especially useful in graph analysis based on a probabilistic framework, wherein the emphasis of this paper also lies. Experiments that take advantage of the possible independence between affinities and costs are left for further work.

We denote by $\mathbf{e}$ the $n \times 1$ vector whose each element is 1.
For an $n\times n$ square matrix $\mathbf{A}$, let $\mathbf{Diag}(\mathbf{A})$ denote the $n \times n$ diagonal matrix whose diagonal elements are the diagonal elements of $\mathbf{A}$ and by $\mathbf{diag}(\mathbf{A})$ the $n \times 1$ vector of the diagonal elements of $\mathbf{A}$. Likewise, for an $n \times 1$ vector $\mathbf{v}$, $\mathbf{Diag}(\mathbf{v})$ denotes the $n \times n$ diagonal matrix containing the elements of vector $\mathbf{v}$ on its diagonal. We use $\exp(\mathbf{A})$ and $\log(\mathbf{A})$ to denote the elementwise exponential and logarithm, respectively; these should not to be confused with the matrix exponential and matrix logarithm which are not used in this article. Furthermore, we use $\mathbf{A} \circ \mathbf{B}$ and $\mathbf{A} \div \mathbf{B}$ for elementwise product and division, respectively, of $n \times m$ matrices $\mathbf{A}$ and $\mathbf{B}$.

%

\section{The shortest path and commute time distances}
\label{sec:DistFams}
The most common distance measure between two nodes of a graph is the \emph{shortest path (SP) distance}. As introduced earlier in Section~\ref{sec:intro}, in our framework, we consider costs associated to the edges of a graph. Hence, we define the SP distance between two nodes as the \emph{minimal cost} of a path between the nodes. This applies for both unweighted and weighted undirected graphs. Also, recall that edge costs can be independent of the affinities $a_{ij}$. Thus, our definition of the SP distance does not necessarily depend on the affinities, either, but only on the costs. In addition, we define the \emph{unweighted SP distance} between two nodes as the \emph{minimal length} of a path between the nodes.\footnote{Some authors, e.g.\ Chebotarev in~\cite{Chebotarev-2011}, instead call the SP distance based on the edge weights the \emph{weighted} SP distance and use the term SP distance only for the distance based on the number of edges on paths. However, there the costs (or resistances) are fixed as the reciprocals of affinities, unlike in our approach.}

The SP distance can be used, for example, for estimating the geodesic distance between points when assuming that the graph points lie on a manifold. One popular method to use this idea is the Isomap algorithm~\cite{Isomap} for nonlinear dimensionality reduction. One major drawback of the SP distance is that it does not take into account the global structure of the network. In particular, it does not consider the number of connections that exist between nodes, only the length of the shortest one.

Another interesting and well-known graph distance measure is the \emph{commute time (CT) distance}~\cite{Gobel-1974} which is defined between two nodes as the \emph{expected length} of paths that a random walker moving along the edges of the graph has to take from one node to the other and back. The transition probability of the walker moving from a node $i$ to an adjacent node $j$ is given naturally as 
\begin{equation}
\label{eq:}
p_{ij}^{\mathrm{ref}} = \dfrac{a_{ij}}{\sum_{k}a_{ik}},
\end{equation}
where the superscript ``ref'' emphasizes that these probabilities will later be considered as reference probabilities, when defining new path probabilities. We refer to a random walk based on these probabilities as a \emph{natural} random walk. The CT distance is well known to be proportional to the \emph{resistance distance}~\cite{Chandra-1989} which is defined as the effective resistance of a network when it is considered as an electric circuit where the poles of a unit volt battery have been attached to the nodes between which the distance is being measured~\cite{Klein-1993, Snell-1984}.

We can also define a generalization of the commute time distance that considers costs of paths instead of their lengths. More precisely, we define the \emph{commute cost (CC) distance} as the \emph{expected cost} of the paths that a random walker will take when moving from a node to another and back according to the transition probabilities $p_{ij}^{\mathrm{ref}}$~\cite{Taylor-1998}. An interesting, somewhat unintuitive result in this context is that \emph{in an undirected graph, the commute cost distance is proportional to the commute time distance}. We provide the proof for this novel result in~\ref{app:CT-CC}. 
Here, it is important to remember that the costs are independent of the weights and vice versa. Thus the same applies between the costs and the reference transition probabilities of the random walker. This result means that the commute time, commute cost and resistance distances are all the same up to a constant factor. In other words, in most practical applications they will give the exact same results, because in practice the interest lies in the ratios of pairwise distances instead of the distances themselves.

A nice thing about the commute time, commute cost and resistance distances, when compared to the SP distance, is that they take into account the number of different paths connecting pairs of nodes. As a result, these distances have been used in different applications of network science with beneficial results. 
However, it has been noted that in a large graph these distances are affected largely by the stationary distribution of the natural random walk on the graph~\cite{Brand-05}. It has recently been shown~\cite{radl2009resistance,vonLuxburg-2010} that in certain models, as the size of a graph grows, the resistance distance (and thus the CT and CC distances as well) between two nodes become only dependent on trivial local properties of the nodes. More specifically, the resistance distance between two nodes approaches the sum of the reciprocals of the degrees of these two nodes. This result is presented in~\cite{vonLuxburg-2010} for random geometric graphs where nodes and edges are added to the graph according to the underlying density. The authors claim that the result can be shown to hold also for other models, such as power law graphs in cases where the minimum degree of the graph grows with the number of nodes.

An intuitive explanation of this phenomenon is that in very large graphs a random walker has too many paths to follow and the chance of the walker finding its destination node becomes more dependent on the number of edges (instead of paths, per se) that lead to that node. This undesirable phenomenon serves as one motivation for defining new graph node distances that choose an alternative between the SP and CT distances. This idea already appeared in the development of the RSP dissimilarity~\cite{RSP, Saerens-2008}, with the main motivations in path planning and simply in proposing a distance interpolating between the SP and CT distances. In the following, we first recapitulate the definition of the RSP dissimilarity and then develop the theory behind it. After this we will review other generalized graph node distances and compare their use in data analysis tasks.

%

\section{Advances in the randomized shortest paths framework}
\label{sec:RSP}
The RSP dissimilarity was defined in~\cite{RSP} inspired by~\cite{Akamatsu-1996} and its theory has been extended further in~\cite{Saerens-2008} and~\cite{Garcia-Diez-2011}. It is based on the interpretation of random walks in terms of statistical physics. The definition involves a parameter\footnote{The authors in the referred work use $\theta$ in place of $\beta$.} $\beta$ which is analogous to the inverse temperature of a thermodynamical system. The RSP dissimilarity is shown to converge to the SP distance as $\beta \rightarrow \infty$ and to the CT distance as $\beta \rightarrow 0^+$.

The reason why the RSP dissimilarity is called a dissimilarity, rather than a distance, is that for intermediate values of the parameter $\beta$, it does not satisfy the triangle inequality, meaning that it is only a semimetric. In the experimental part of the paper, we focus on the effect of the choice of a distance measure on clustering and classification algorithms. 
When studying such algorithms, it is often assumed that they are used in conjunction with a metric, i.e.,
a distance measure that satisfies the triangle inequality. 
Also, the triangle inequality can be exploited in order to improve the efficiency of some distance-based algorithms, cf.~\cite{Elkan2003}.
However, in our study, we use a kernel $k$-means algorithm for clustering and a simple 1-nearest-neighbor algorithm for semi-supervised classification, which both can be used even with a semimetric. Furthermore, it has been already shown that using the RSP dissimilarity with its intermediate parameter values provides good results in graph node clustering~\cite{RSP}.

\subsection{The randomized shortest path dissimilarity}
\label{sec:RSP-def}
We shall first recall the definition of the RSP dissimilarity~\cite{RSP, Saerens-2008}. It is defined by considering a random walker choosing an \emph{absorbing}, or \emph{hitting}, path from a starting node $s$ to a destination node $t$, meaning that node $t$ can appear in the path only once, as the ending node. Let $\mathcal{P}_{st}$ denote the set of such paths and let $\wp = (i_{1} = s,\ldots, i_{T} = t) \in \mathcal{P}_{st}$. The \emph{reference probability} of the path $\wp$ is $\widetilde{\mathrm{P}}_{st}^{\mathrm{ref}}(\wp) = p_{i_{1}i_{2}}^{\mathrm{ref}} \cdots p_{i_{T-1} i_{T}}^{\mathrm{ref}}$. It simply corresponds to the likelihood of the paths, i.e., the product of the transition probabilities along the path. 

In the RSP model, the randomness of the walker is constrained by fixing the relative entropy between the distribution over paths according to the reference probabilities and the distribution over paths that the walker actually chooses from. With this constraint, the walker then chooses the path from the probability distribution that minimizes the \emph{expected cost}\footnote{Notice the difference in notation between the expected cost (denoted with a bar as $\overline{c}$) and the cost of a particular path (denoted with a tilde as $\widetilde{c}$).}
\[
\overline{c}(\widetilde{\mathrm{P}}_{st})
= \! \sum_{\wp \in \mathcal{P}_{st}}\! \widetilde{\mathrm{P}}_{st}(\wp)\widetilde{c}(\wp)
\]
of going from node $s$ to node $t$. Thus, the relative entropy constraint controls the \emph{exploration} of the walker, whereas the minimization of expected cost controls its \emph{exploitation}. 
Formally, the walker moves according to the distribution
\begin{equation}
\label{eq:RSPmin}
\widetilde{\mathrm{P}}_{st}^{\mathrm{RSP}} = \argmin_{\widetilde{\mathrm{P}}_{st}} \overline{c}(\widetilde{\mathrm{P}}_{st}) 
\ \ 
\textrm{subject to}
\ \ 
\begin{cases}
J(\widetilde{\mathrm{P}}_{st} \| \widetilde{\mathrm{P}}_{st}^{\mathrm{ref}}) = J_{0} \\ 
\sum\limits_{\wp \in \mathcal{P}_{st}}\widetilde{\mathrm{P}}_{st}(\wp) = 1
\end{cases}
\end{equation}
where $J(\widetilde{\mathrm{P}}_{st} \| \widetilde{\mathrm{P}}_{st}^{\mathrm{ref}}) = \sum_{\wp \in \mathcal{P}_{st}} \! \widetilde{\mathrm{P}}_{st}(\wp) \log (\widetilde{\mathrm{P}}_{st}(\wp) / \widetilde{\mathrm{P}}_{st}^{\mathrm{ref}}(\wp))$ is the relative entropy of the distribution with respect to $\widetilde{\mathrm{P}}^{\mathrm{ref}}$, which is constrained to a value $J_{0}$.
The minimization is shown~\cite{RSP,Saerens-2008} to result in a Boltzmann distribution
\begin{equation}
\label{eq:RSPprob}
\widetilde{\mathrm{P}}_{st}^{\mathrm{RSP}}(\wp) = \dfrac{\widetilde{\mathrm{P}}_{st}^{\mathrm{ref}}(\wp) \exp \left( -\beta \widetilde{c}(\wp)\right)}{\sum\limits_{\wp \in \mathcal{P}_{st}}\widetilde{\mathrm{P}}_{st}^{\mathrm{ref}}(\wp) \exp \left( -\beta \widetilde{c}(\wp)\right)},
\end{equation}
where the inverse temperature parameter $\beta$ controls the influence of the cost on the walker's selection of a path. When applying the model, the user is assumed to provide $\beta$ as an input parameter instead of the relative entropy $J_{0}$. 

After deriving the optimal distribution for minimizing the expected cost, the RSP dissimilarity between the nodes $s$ and $t$ is defined as this expected cost after symmetrization, formally
\begin{equation}
\dist_{st}^{\mathrm{RSP}} = \left(\overline{c}(\widetilde{\mathrm{P}}_{st}^{\mathrm{RSP}}) + \overline{c}(\widetilde{\mathrm{P}}_{ts}^{\mathrm{RSP}}) \right)/2.
\end{equation}
The authors in~\cite{RSP} develop an algorithm for computing the expected cost $\overline{c}(\widetilde{\mathrm{P}}_{st}^{\mathrm{RSP}})$, which is not at all a trivial task. In the next section we develop a new, more efficient algorithm for computing the expected costs and thus the matrix $\mathbf{\Delta}^{\mathrm{RSP}}$ of the RSP dissimilarities between all pairs of nodes of a graph in closed form. After that, in Section~\ref{sec:FE-dist}, we will consider a distance based on the minimized \emph{Helmholtz free energy}~\cite{Peliti-2011}, instead of minimized expected cost. The free energy of a thermodynamical system with temperature $T = 1/\beta$ and state transition probabilities $\widetilde{\mathrm{P}}_{st}$ has the form\footnote{Conventionally, the Helmholtz free energy is defined with the entropy of $\widetilde{\mathrm{P}}_{st}$ in place of the relative entropy. Regardless of this, we use the term as we have presented it.}
\begin{equation}
\phi(\widetilde{\mathrm{P}}_{st}) = \overline{c}(\widetilde{\mathrm{P}}_{st}) + J(\widetilde{\mathrm{P}}_{st} \| \widetilde{\mathrm{P}}_{st}^{\mathrm{ref}})/\beta.
\label{eq:FEdef}
\end{equation}


\subsection{Algorithm for faster computation of $\mathbf{\Delta}^{\mathrm{RSP}}$}
\label{sec:RSPbatch}
We now show how to compute the RSP dissimilarity and then develop an algorithm that allows computing the set of all pairwise RSP dissimilarities between the nodes of a graph in a batch mode. The algorithm in the original reference~\cite{RSP} performs a loop over all the nodes of the graph and computes the needed quantities considering one node as the destination node at a time. Our new algorithm is based solely on matrix manipulations and can thus provide faster execution than a na\"ive looping.

The computation of the expected cost $\overline{c}(\widetilde{\mathrm{P}}_{st}^{\mathrm{RSP}})$ is based on considering the denominator of the right side of Equation~(\ref{eq:RSPprob}) and denoting
\begin{equation}
\label{eq:PartitionFunction}
z^{\mathrm{h}}_{st} = \sum_{\wp \in \mathcal{P}_{st}} \widetilde{\mathrm{P}}_{st}^{\mathrm{ref}}(\wp) \exp (-\beta\widetilde{c}(\wp)). 
\end{equation}
This quantity is in statistical physics called the \emph{partition function} of a thermodynamical system. In our case, the system consists of the hitting (hence the superscript ``h'' in $z^{\mathrm{h}}_{st}$) paths of $\mathcal{P}_{st}$. The partition function is essential for deriving different quantities related to the RSP framework. 
Indeed, by manipulating the expected cost of travelling from node $s$ to node $t$ we see that
\begin{align}
\label{eq:ExpCost}
\overline{c}(\widetilde{\mathrm{P}}_{st}^{\mathrm{RSP}})
&= \sum_{\wp \in \mathcal{P}_{st}} \widetilde{\mathrm{P}}_{st}^{\mathrm{RSP}}(\wp)\widetilde{c}(\wp) 
= \dfrac{\sum_{\wp \in \mathcal{P}_{st}} \widetilde{\mathrm{P}}_{st}^{\mathrm{ref}}(\wp) \exp(-\beta\widetilde{c}(\wp)) \widetilde{c}(\wp)}{z^{\mathrm{h}}_{st}} \nonumber \\
&= -\dfrac{1}{z^{\mathrm{h}}_{st}} \dfrac{\partial z^{\mathrm{h}}_{st}}{\partial \beta} 
= -\dfrac{\partial \log z^{\mathrm{h}}_{st}}{\partial \beta}
\end{align}
meaning that the expected cost can be obtained by taking the derivative of the logarithm of the partition function.

Let us denote by $\mathbf{C}$ the matrix of costs on edges, $c_{ij}$, and by $\mathbf{P}^{\mathrm{ref}}$ the transition probability matrix of the natural random walk associated to the graph $G$ containing the elements $p_{ij}^{\mathrm{ref}}$. The latter can be computed from the adjacency matrix as $\mathbf{P}^{\mathrm{ref}} = \mathbf{D}^{-1} \mathbf{A}$, where $\mathbf{D} = \mathbf{Diag}(\mathbf{Ae})$. In order to compute the partition function, we define a new matrix
\begin{equation}
\label{eq:W}
\mathbf{W} = \mathbf{P^{\mathrm{ref}}} \circ \exp (-\beta \mathbf{C}).
\end{equation}
This matrix is substochastic and thus it can be interpreted as a new transition matrix defining a \emph{killed}, or an \emph{evaporating} random walk on the graph~\cite{Fouss-2012}. This means that at each step of the walk the random walker has a non-zero probability of stopping its walk, i.e.\ getting killed. Another way of interpreting this is by imagining an additional, absorbing ``cemetery'' node~\cite{Bavaud-2012}, where the walker can end up from each node of the graph with a non-zero probability.

Remember now that we only consider hitting paths, meaning that we want to set the destination node $t$ absorbing. For this, we define a new matrix by setting the row $t$ of $\mathbf{W}$ to zero: $\mathbf{W}_{t}^{\mathrm{h}} = \mathbf{W} - \mathbf{e}_{t} (\mathbf{w}_{t}^{\mathrm{r}})^{\mathsf{T}}$, where $\mathbf{e}_{t}$ is a vector containing 1 in element $t$ and 0 elsewhere and $\mathbf{w}_{t}^{\mathrm{r}}$ is row $t$ (hence the superscript ``r'') of matrix $\mathbf{W}$ as a column vector, i.e.\ $\mathbf{w}_{t}^{\mathrm{r}} = (\mathbf{e}_{t}^{\mathsf{T}} \mathbf{W})^{\mathsf{T}}$. Thus, expressed elementwise, $(\mathbf{W}_{t}^{\mathrm{h}})_{ij} = (\mathbf{W})_{ij}$, if $i \neq t$, and 0 otherwise. 

The powers of this matrix, $(\mathbf{W}_{t}^{\mathrm{h}})^{\tau}$, contain in element $(s,t)$ the probability that a killed random walk of exactly $\tau$ steps leaving from node $s$ ends up in node $t$ when obeying the transition probabilities assigned by $\mathbf{W}_{t}^{\mathrm{h}}$. This can also be expressed as
\[
\left[(\mathbf{W}_{t}^{\mathrm{h}})^{\tau}\right]_{st} = \sum_{\wp \in \mathcal{P}_{st}(\tau)} \widetilde{\mathrm{P}}_{st}^{\mathrm{ref}}(\wp) \exp(-\beta \widetilde{c}(\wp)),
\]
where $\mathcal{P}_{st}(\tau)$ denotes the set of paths whose length is $\tau$ going from node $s$ to node $t$. Then by summing over all walk lengths\!\!
\footnote{Although in~\cite{RSP} only paths of length $\geq 1$ are considered, we also include paths of length 0, i.e.\ the paths that consist of only one node and no links, into the set of allowed paths; see~\cite{BoP} for a discussion related to this.}
$\tau$ we can cover all hitting paths from node $s$ to $t$ and write the partition function as a power series
\begin{align*}
z^{\mathrm{h}}_{st}
&= \sum_{\wp \in \mathcal{P}_{st}} \widetilde{\mathrm{P}}_{st}^{\mathrm{ref}}(\wp) \exp(-\beta\widetilde{c}(\wp))
= \sum_{\tau = 0}^{\infty} \! \sum_{\wp \in \mathcal{P}_{st}(\tau)} \widetilde{\mathrm{P}}_{st}^{\mathrm{ref}}(\wp) \exp(-\beta \widetilde{c}(\wp)) \\
&= \sum_{\tau = 0}^{\infty} \left[(\mathbf{W}_{t}^{\mathrm{h}})^{\tau}\right]_{st}
= \left[(\mathbf{I} - \mathbf{W}_{t}^{\mathrm{h}})^{-1}\right]_{st},
\end{align*}
The series converges to the matrix $\mathbf{Z}_{t}^{\mathrm{h}} = (\mathbf{I} - \mathbf{W}_{t}^{\mathrm{h}})^{-1}$, as the spectral radius of $\mathbf{W}_{t}^{\mathrm{h}}$ is less than one, $\rho(\mathbf{W}_{t}^{\mathrm{h}}) < 1$. The matrix $\mathbf{Z}_{t}^{\mathrm{h}}$ is the \emph{fundamental matrix} corresponding to the killed Markov Chain with the transition matrix $\mathbf{W}_{t}^{\mathrm{h}}$. In the original reference~\cite{RSP}, the authors then derive a way of computing the matrix $\mathbf{Z}_{t}^{\mathrm{h}}$ just by using the simpler matrix
\begin{equation}
\label{eq:Z}
\mathbf{Z} = (\mathbf{I}-\mathbf{W})^{-1}.
\end{equation}
This is the fundamental matrix of the killed, but \emph{non-absorbing} Markov Chain with the transition matrix $\mathbf{W}$, and its computation only requires a simple matrix inversion.

With this in mind, the authors in~\cite{RSP} use the Sherman-Morrison rule for a rank-one update of a matrix 
for deriving the form (Equation (20) in~\cite{RSP})
\begin{equation}
\label{eq:Zth}
\mathbf{Z}_{t}^{\mathrm{h}} = \mathbf{Z} - \dfrac{\mathbf{Z} \mathbf{e}_{t} (\mathbf{w}_{t}^{\mathrm{r}})^{\mathsf{T}} \mathbf{Z}}{1 + (\mathbf{w}_{t}^{\mathrm{r}})^{\mathsf{T}} \mathbf{Z} \mathbf{e}_{t}}.
\end{equation}
They then use this form of $\mathbf{Z}_{t}^{\mathrm{h}}$ for computing a vector of dissimilarities from all nodes of the graph to one fixed absorbing destination node $t$ at a time. In order to compute all dissimilarities in the graph, the algorithm performs a loop over $t$ across all the nodes of the graph considering them as absorbing one at a time.

However, note that we are only interested in computing the element $(s,t)$ of the matrix $\mathbf{Z}_{t}^{\mathrm{h}}$, which we can do by using~(\ref{eq:Zth}):
\begin{equation}
\label{eq:zsth}
\begin{aligned}
z_{st}^{\mathrm{h}} 
&= \mathbf{e}_{s}^{\mathsf{T}} \mathbf{Z}_{t}^{\mathrm{h}} \mathbf{e}_{t} 
 = z_{st} - \dfrac{    z_{st}(\mathbf{w}_{t}^{\mathrm{r}})^{\mathsf{T}} \mathbf{Z} \mathbf{e}_{t}    }{    1 + (\mathbf{w}_{t}^{\mathrm{r}})^{\mathsf{T}} \mathbf{Z} \mathbf{e}_{t}    }
 = z_{st}\left(    1 - \dfrac{    1 + (\mathbf{w}_{t}^{\mathrm{r}})^{\mathsf{T}} \mathbf{Z} \mathbf{e}_{t} - 1    }{  1 + (\mathbf{w}_{t}^{\mathrm{r}})^{\mathsf{T}} \mathbf{Z} \mathbf{e}_{t}  }    \right) \\
&= \dfrac{z_{st}}{ \mathbf{e}_{t}^{\mathsf{T}} \mathbf{e}_{t} + \mathbf{e}_{t}^{\mathsf{T}}\mathbf{W} \mathbf{Z} \mathbf{e}_{t}}
 = \dfrac{z_{st}}{ \mathbf{e}_{t}^{\mathsf{T}}\left( \mathbf{I} + \mathbf{WZ} \right) \mathbf{e}_{t} }
 = \dfrac{z_{st}}{\mathbf{e}_{t}^{\mathsf{T}} \mathbf{Z} \mathbf{e}_{t}}
 = \dfrac{z_{st}}{z_{tt}},
\end{aligned}
\end{equation}
where $\mathbf{Z} = \mathbf{I} + \mathbf{WZ}$ follows from $(\mathbf{I} - \mathbf{W})\mathbf{Z} = \mathbf{I}$.

Thus, we can compute the matrix $\mathbf{Z}^{\mathrm{h}}$, whose element $(s,t)$ is the partition function $z_{st}^{\mathrm{h}}$ of hitting paths from node $s$ to node $t$ without having to compute $\mathbf{Z}_{t}^{\mathrm{h}}$ for each $t$ separately. This can be done simply using the matrix $\mathbf{Z}$ as
\begin{equation}
\mathbf{Z}^{\mathrm{h}} = \mathbf{Z}\mathbf{D}_{\mathrm{h}}^{-1},
\label{eq:zHit}
\end{equation}
where $\mathbf{D}_{\mathrm{h}} = \mathbf{Diag}(\mathbf{Z})$ is the diagonal matrix with elements $z_{ii}$ on its diagonal. The advantage compared to the original algorithm presented in~\cite{RSP} is that $\mathbf{Z}^{\mathrm{h}}$ can be used for computing the RSP dissimilarities between all pairs of nodes instead of having to compute the matrix $\mathbf{Z}_{t}^{\mathrm{h}}$ separately for each destination $t$. This reduces the computational complexity of the algorithm and makes it much more straightforward to implement than before.

Note that a similar derivation for computing $z_{st}^{h}$ as the above one is also presented in Appendix A of~\cite{BoP}. It is also showed in the same reference that, in fact, the partition function in this context gives the probability that a random walker starting from $s$ using the transition matrix $\mathbf{W}_{t}^{\mathrm{h}}$ will reach $t$ before getting killed, i.e.\ before ending in the imaginary cemetery node.

Now we can finally derive the new matrix formula for the RSP dissimilarities between all pairs of nodes using Equations (\ref{eq:ExpCost}) and (\ref{eq:zHit}). The expected cost is given by
\begin{equation}
\label{eq:ExpCost2}
\overline{c}(\widetilde{\mathrm{P}}_{st}^{\mathrm{RSP}}) = -\dfrac{\partial \log z^{\mathrm{h}}_{st}}{\partial \beta} = -\dfrac{\partial \log (z_{st} / z_{tt})}{\partial \beta} = -\dfrac{\partial \log z_{st}}{\partial \beta} + \dfrac{\partial \log z_{tt}}{\partial \beta}.
\end{equation}
The first term can be computed by
\begin{align*}
\dfrac{\partial \log z_{st}}{\partial \beta}
& = \dfrac{1}{z_{st}} \dfrac{\partial z_{st}}{\partial \beta} 
  = \dfrac{1}{z_{st}} \dfrac{\partial \mathbf{e}_{s}^{\mathsf{T}} \mathbf{Z} \mathbf{e}_{t}}{\partial \beta}
  = \dfrac{1}{z_{st}} \mathbf{e}_{s}^{\mathsf{T}} \dfrac{\partial (\mathbf{I} - \mathbf{W})^{-1} }{\partial \beta} \mathbf{e}_{t} \\
& = -\dfrac{1}{z_{st}} \mathbf{e}_{s}^{\mathsf{T}} (\mathbf{I} - \mathbf{W})^{-1} \dfrac{\partial (\mathbf{I} - \mathbf{W}) }{\partial \beta} (\mathbf{I} - \mathbf{W})^{-1} \mathbf{e}_{t} \\
& = \dfrac{1}{z_{st}} \mathbf{e}_{s}^{\mathsf{T}} \mathbf{Z} \dfrac{\partial \mathbf{W} }{\partial \beta} \mathbf{Z} \mathbf{e}_{t} \\
& = -\dfrac{1}{z_{st}} \mathbf{e}_{s}^{\mathsf{T}} \mathbf{Z} \left( \mathbf{C} \circ \mathbf{W} \right) \mathbf{Z} \mathbf{e}_{t},
\end{align*}
where we used $\dfrac{\partial \mathbf{W} }{\partial \beta} = \dfrac{\partial}{\partial \beta} \left(\mathbf{P}_{st}^{\mathrm{ref}} \circ \exp( -\beta \mathbf{C}) \right) = -\left(\mathbf{C} \circ \mathbf{W} \right)$.

Thus, we can write Equation~(\ref{eq:ExpCost2}) as
\begin{equation}
\label{eq:RSPclosedIntermediate}
\overline{c}(\widetilde{\mathrm{P}}_{st}^{\mathrm{RSP}}) = \dfrac{\mathbf{e}_{s}^{\mathsf{T}} \mathbf{Z} \left( \mathbf{C} \circ \mathbf{W} \right) \mathbf{Z} \mathbf{e}_{t}}{z_{st}}
- \dfrac{\mathbf{e}_{t}^{\mathsf{T}} \mathbf{Z} \left( \mathbf{C} \circ \mathbf{W} \right) \mathbf{Z} \mathbf{e}_{t}}{z_{tt}}
\end{equation}
Let us then denote $\mathbf{S} = \left(\mathbf{Z}(\mathbf{C} \circ \mathbf{W}) \mathbf{Z}\right) \div \mathbf{Z}$, where $\div$ marks elementwise division.
In fact, $\mathbf{S}$ is the matrix form of the first term on the right side of Equation~(\ref{eq:RSPclosedIntermediate}),
and contains the expected costs of non-hitting random walks. We can now use it to write out the matrix form of computing all the expected costs of hitting walks as
\[
\overline{\mathbf{C}}
= \mathbf{S} - \mathbf{e} \mathbf{d}_{\mathbf{S}}^{\mathsf{T}},
\]
where $\mathbf{d}_{\mathbf{S}} = \mathbf{diag}(\mathbf{S})$ is the vector of diagonal elements of $\mathbf{S}$.
Finally, the matrix of RSP dissimilarities $\mathbf{\Delta}^{\mathrm{RSP}}$ is defined by symmetrizing $\overline{\mathbf{C}}$:
$\mathbf{\Delta}^{\mathrm{RSP}} = ( \overline{\mathbf{C}} + \overline{\mathbf{C}}^{\mathsf{T}} ) /2$. 

\subsection{A new generalized distance based on Helmholtz free energy}
\label{sec:FE-dist}
As already mentioned earlier, one of the drawbacks of the RSP dissimilarity is that it is not a metric as it does not necessarily satisfy the triangle inequality for intermediate values of $\beta$.
To overcome this problem we derive a new distance measure
called the \emph{free energy (FE) distance}, which is based on the same idea behind the RSP dissimilarity.
We conclude that the proposed FE distance is actually the same as the \emph{potential distance} defined recently in Equation (38) of~\cite{BoP} based on a \emph{bag-of-paths framework}. However, in that reference, the derivation of the potential distance is left rather unmotivated. The derivation provided here gives a more sound theoretical background to the distance measure and thus we suggest to call the distance the free energy distance instead of the potential distance. 

Free energy has already been used in various contexts in network theory. In~\cite{Delvenne-2011}, the authors define a ranking method called the free-energy rank (in the spirit of the well-known PageRank~\cite{Page-1998}) by computing the transition probabilities minimizing the free energy rate encountered by a random walker. Then, the stationary distribution of the defined Markov chain is the free-energy rank score. In~\cite{Bavaud-2012}, the authors compute edge flows minimizing the free energy between two nodes. The resulting flows define some new edge and node betweenness measures, balancing exploration and exploitation through an adjustable temperature parameter. Their model is quite close to the RSP framework and was developed parallel to our article. However, the authors do not define a distance measure based on the free energy.

We now derive the FE distance and then show that it coincides with the potential distance. Recall that the RSP dissimilarity was defined by considering a distribution of random walks between two nodes that minimizes the expected cost $\overline{c}(\widetilde{\mathrm{P}}_{st})$ subject to a relative entropy constraint. Now, instead of the expected cost, let us consider a random walker choosing a path from node $s$ to node $t$ according to the distribution that minimizes the free energy according to Equation~(\ref{eq:FEdef}). The minimization of free energy can be simply written as
\begin{align*}
\widetilde{\mathrm{P}}_{st}^{\mathrm{FE}}
= &\argmin_{\widetilde{\mathrm{P}}_{st}} \! \sum_{\wp \in \mathcal{P}_{st}} \! \widetilde{\mathrm{P}}_{st}(\wp)\widetilde{c}(\wp)
+\frac{1}{\beta} \! \sum_{\wp \in \mathcal{P}_{st}} \! \widetilde{\mathrm{P}}_{st}(\wp) \log (\widetilde{\mathrm{P}}_{st}(\wp) / \widetilde{\mathrm{P}}_{st}^{\mathrm{ref}}(\wp)),
\\ 
&\textrm{subject to}
\sum\limits_{\wp \in \mathcal{P}_{st}}\widetilde{\mathrm{P}}_{st}(\wp) = 1.
\end{align*}
It is not difficult to see that this problem becomes equivalent to the minimization problem~(\ref{eq:RSPmin}) involved in the definition of the RSP probabilities and thus the optimal solution is again the Boltzmann distribution~(\ref{eq:RSPprob}), in other words, $\widetilde{\mathrm{P}}_{st}^{\mathrm{FE}} = \widetilde{\mathrm{P}}_{st}^{\mathrm{RSP}}$. We define the free energy distance between nodes $s$ and $t$ as the symmetrized minimum free energy between these two nodes, in other words
\begin{equation}
\label{eq:FEdist}
\dist_{st}^{\mathrm{FE}} = \left( \phi(\widetilde{\mathrm{P}}_{st}^{\mathrm{FE}}) + \phi(\widetilde{\mathrm{P}}_{ts}^{\mathrm{FE}}) \right) /2
\end{equation}


In order to show that the FE distance coincides with the potential distance defined in Equation (38) of~\cite{BoP}, we remind that the FE probability (which is equal to the RSP probability) of a path $\wp$ can be written as (see Equations (\ref{eq:RSPprob}) and (\ref{eq:PartitionFunction}))
\[
\widetilde{\mathrm{P}}_{st}^{\mathrm{FE}}(\wp)
= \dfrac{\widetilde{\mathrm{P}}_{st}^{\mathrm{ref}}(\wp) \exp (-\beta \widetilde{c}(\wp))}{z^{\mathrm{h}}_{st}}.
\]
Using then the fact that $\sum_{\wp \in \mathcal{P}_{st}} \negthickspace \widetilde{\mathrm{P}}_{st}^{\mathrm{FE}}(\wp) = 1$, we can write out the expression for the relative entropy:
\begin{align*}
& J(\widetilde{\mathrm{P}}_{st}^{\mathrm{FE}} \| \widetilde{\mathrm{P}}_{st}^{\mathrm{ref}}) 
= \sum_{\wp \in \mathcal{P}_{st}} \! \widetilde{\mathrm{P}}_{st}^{\mathrm{FE}}(\wp)
\log \left(\widetilde{\mathrm{P}}_{st}^{\mathrm{FE}}(\wp) / \widetilde{\mathrm{P}}_{st}^{\mathrm{ref}}(\wp) \right) \\
& = \sum_{\wp \in \mathcal{P}_{st}} \! \widetilde{\mathrm{P}}_{st}^{\mathrm{FE}}(\wp) \log \left( \dfrac{\widetilde{\mathrm{P}}_{st}^{\mathrm{ref}}(\wp) \exp (-\beta \widetilde{c}(\wp))}{z^{\mathrm{h}}_{st}} \right)
- \sum_{\wp \in \mathcal{P}_{st}} \widetilde{\mathrm{P}}_{st}^{\mathrm{FE}}(\wp) \log \left( \widetilde{\mathrm{P}}_{st}^{\mathrm{ref}}(\wp) \right) \\
& =  \sum_{\wp \in \mathcal{P}_{st}} \! \widetilde{\mathrm{P}}_{st}^{\mathrm{FE}}(\wp) \log \left( \widetilde{\mathrm{P}}_{st}^{\mathrm{ref}}(\wp) \right)
- \beta \! \! \! \sum_{\wp \in \mathcal{P}_{st}} \! \widetilde{\mathrm{P}}_{st}^{\mathrm{FE}}(\wp)\widetilde{c}(\wp) \\
& \quad - \log(z^{\mathrm{h}}_{st})
- \! \! \! \sum_{\wp \in \mathcal{P}_{st}} \! \widetilde{\mathrm{P}}_{st}^{\mathrm{FE}}(\wp) \log \left( \widetilde{\mathrm{P}}_{st}^{\mathrm{ref}}(\wp) \right) \\ 
& =  -\beta \overline{c}(\widetilde{\mathrm{P}}_{st}^{\mathrm{FE}}) - \log(z^{\mathrm{h}}_{st})
\end{align*}
When combining this result with Equation~(\ref{eq:FEdef}), the FE becomes
\[
\phi(\widetilde{\mathrm{P}}_{st}^{\mathrm{FE}}) = -\dfrac{1}{\beta}\log(z^{\mathrm{h}}_{st})
\]
which after the symmetrization~(\ref{eq:FEdist}) equals the potential distance defined in Equation (38) of~\cite{BoP}. Thanks to the derivation of the new algorithm for the RSP dissimilarities in Section~\ref{sec:RSPbatch}, the matrix of free energies between all pairs of nodes
\[
\mathbf{\Phi} = -1/\beta \log{\mathbf{Z}^{\mathrm{h}}}
\]
can also be computed straightforwardly by performing Equations~(\ref{eq:W}), (\ref{eq:Z}) and (\ref{eq:zHit}). Finally, the matrix of all FE distances is obtained, again, by symmetrization as $\mathbf{\Delta}^{\mathrm{FE}} = (\mathbf{\Phi} + \mathbf{\Phi}^{\mathsf{T}})/2$.

Thus, we have shown that the potential distance derived within the bag-of-paths framework in~\cite{BoP}, in fact can be derived from the RSP framework by considering the minimum Helmholtz free energy as the distance, instead of the minimum expected cost. The minimization of the FE can be interpreted as a regularized version of the RSP model where the random walker finds a compromise between the expected cost and predictability of its path from $s$ to $t$. The compromise is controlled by the inverse temperature $\beta$.

Of course, the FE distance also satisfies all the properties that were proved for the potential distance in~\cite{BoP}. Most importantly, it was shown that the distance obeys the triangle inequality (in Inequality (37) of~\cite{BoP}) and is thus a metric, as opposed to the RSP dissimilarity. The distance also converges to the SP distance when $\beta \to \infty$ and to the CT distance\footnote{To the CT distance divided by 2, to be precise.} when $\beta \to 0^{+}$ (see Appendices D and E in~\cite{BoP}). In addition, in Appendix C of~\cite{BoP}, it is shown to be \emph{graph geodetic}~\cite{Chebotarev-2011,Klein-1998}, meaning that $\dist_{st}^{\mathrm{FE}} = \dist_{sk}^{\mathrm{FE}} + \dist_{kt}^{\mathrm{FE}}$ if and only if all paths from node $s$ to node $t$ go through node $k$. This shows that the minimum free energy between two nodes defines a meaningful distance measure between graph nodes with nice properties. Interestingly, as can be seen from Equation~(\ref{eq:FEdef}), the FE distance is actually a metric resulting from the sum of two dissimilarities, namely the expected cost (i.e.\ the RSP dissimilarity, after symmetrization) and the relative entropy, neither of which satisfy the triangle inequality by themselves. We also note that the quantity $\log z_{st}^{\mathrm{h}}$ already appeared in~\cite{Garcia-Diez-2011} as a potential inducing a drift for a random walker in a continuous-state extension of the RSP framework.

\section{Related work on generalized graph distances}
\label{sec:OtherDists}
There have been a few other suggestions for graph distances that generalize the resistance or CT and the SP distances, which we will discuss in this section. Of course, the simplest interpolation between the two distances is their weighted average, which we will experiment with as a null model.
In addition, Chebotarev has defined several parametrized graph distance measures~\cite{Chebotarev-2011, Chebotarev-1997, Chebotarev-2002, Chebotarev-2012}. In this paper, we focus on the logarithmic forest distances~\cite{Chebotarev-2011}.
Alamgir and von~Luxburg defined a generalized distance called the \emph{$p$-resistance distance} in order to tackle the problem of the resistance distance becoming meaningless with large graphs~\cite{pRes}. Indeed they show that with certain values of the parameter $p$, the $p$-resistance distance avoids this pitfall.

\subsection{The SP-CT combination distance}
\label{sec:SP-CT}
The graph distance families presented and reviewed in this paper all involve a sophisticated theoretical derivation. In the experiments we want to compare these distances also to a baseline model that generalizes the SP and CT distances in the most simple way, namely the weighted average of the two distances:
\begin{equation}
\dist_{st}^{\mathrm{SP-CT}} = \lambda \dist_{st}^{\mathrm{SP}} + (1-\lambda) \dist_{st}^{\mathrm{CT}},
\end{equation}
where $\lambda \in [0,1]$.\footnote{In the experiments, we actually use a linear combination of the SP distance and resistance distance. This is because the CT distance values tend to be very large compared to the SP distances, whereas resistance distance values are in the same magnitude as SP distances.} We call it straightforwardly the SP-CT combination distance; it satisfies the triangle inequality because a convex combination of metrics is also a metric. Although the SP-CT combination does not contain as interesting details as the other distances, we can see that in some cases even using this simple choice can be competitive with the other parametrized distances.

\subsection{Logarithmic forest distances} 
The logarithmic forest distance~\cite{Chebotarev-2011} has its foundation in the matrix-forest theorem and another family of distances developed earlier by Chebotarev called simply the forest distance~\cite{Chebotarev-1997, Chebotarev-2002}.
The definition of the logarithmic forest distance goes as follows. First, we define the Laplacian (or Kirchhoff) matrix of a graph $G$ as $\mathbf{L} = \mathbf{D} - \mathbf{A}$, where $\mathbf{D} = \mathbf{Diag}(\mathbf{Ae})$. Then we consider the matrix
\[
\mathbf{Q} = (\mathbf{I} + \alpha \mathbf{L})^{-1},
\]
where $\alpha > 0$. The elements of this matrix measure the \emph{relative forest accessibilities}~\cite{Chebotarev-1997} which can be considered as similarities between nodes of the graph after all its edge weights have been multiplied by the constant $\alpha$. In fact, in~\cite{Chebotarev-2011} Chebotarev handles a more general case by considering arbitrary transformations of the edge weights and multigraphs instead of graphs. The definition proceeds by taking the elementwise logarithmic transformation
\[
\mathbf{M} = \gamma (\alpha -1) \log_{\alpha}\!\mathbf{Q},
\]
where $\gamma > 0$ is another parameter and the logarithm is taken elementwise in basis $\alpha$. This expression provides another similarity measure. From it, the matrix of logarithmic forest distances is derived as
\begin{equation}
\mathbf{\Delta}^{\mathrm{logFor}} = \frac{1}{2}(\mathbf{m}\mathbf{e}^{\mathsf{T}} + \mathbf{e}\mathbf{m}^{\mathsf{T}}) - \mathbf{M},
\end{equation}
where $\mathbf{m} = \mathbf{diag}(\mathbf{M})$. The last transition is a classical way of defining a matrix of distances from a matrix of similarities~\cite{ModernMDS}.


The definition provides a metric which also satisfies the graph-geodetic property (see Section~\ref{sec:FE-dist}). It involves two parameters $\gamma$ and $\alpha$. For any positive value of the parameter $\gamma$, the logarithmic forest distance becomes proportional to the CT and the SP distances as $\alpha \to 0^{+}$ and $\alpha \to \infty$, respectively\footnote{More accurately, the logarithmic forest distance converges to the \emph{unweighted} SP distance (see Section~\ref{sec:DistFams}) but we nevertheless include it in our comparison.}. In the special case of $\gamma = \log(e + \alpha^{2/n})$, Chebotarev shows that the logarithmic forest distance approaches exactly these two other distances. However, this form is not very practical, because even with moderate size graphs the exponent $2/n$ cancels out the effect of setting a large value to $\alpha$. Thus, we decided simply to assign $\gamma = 1$ in our experiments.

\subsection{\textit{p}-resistance distance}
\label{sec:pRes}
Alamgir and von~Luxburg~\cite{pRes} defined a generalization of the resistance distance, called the \emph{$p$-resistance distance}. 
Like the resistance distance, the $p$-resistance distance considers the graph as an electrical resistance network, where the edges $(k,l) \in E$ of the network have resistances $r_{kl}$ and a unit volt battery is attached to the target nodes whose distance is being measured. This forms a \emph{unit flow from $s$ to $t$}, $(i_{kl})_{s \to t}$, where the currents $i_{kl}$ are assigned on all the edges $(k,l) \in E$ of the graph. In short, this means that for all $k,l$ such that $(k,l)\in E$ the currents $i_{kl}$ satisfy the following three conditions: (1) $i_{kl} = -i_{lk}$, (2) $\sum_{l} i_{sl} = 1$ and $\sum_{k} i_{kt} = -1$ and (3) $\sum_{l} i_{kl} = 0$ for $s \neq k \neq t$. Then for a constant $p > 0$, the $p$-resistance distance is defined as the minimized $p$-resistance (w.r.t.\ the unit flow) between $s$ and $t$, formally as
\begin{equation}
\label{eq:pRes}
\dist_{st}^{p\mathrm{Res}} = \min_{(i_{kl})_{s \to t}}\!\left\{ \sum_{(k,l) \in E} r_{kl}|i_{kl}|^{p} \ \Big| \ (i_{kl})_{s \to t} \textrm{ is a unit flow from \textit{s} to \textit{t}} \right\}\!.
\end{equation}
When the parameter $p = 2$, the above definition becomes the definition of effective resistance, i.e.\ the resistance distance and when $p = 1$ the distance coincides with the SP distance. Alamgir and Von Luxburg~\cite{pRes} show that there exists a value $1<p<2$ for which the $p$-resistance distance avoids the problem of the traditional resistance distance with large graphs, introduced in Section~\ref{sec:DistFams}.


Another, almost identical form of the $p$-resistance distance has been proposed by Herbster in~\cite{Herbster-2010}. Also, in a closely related work~\cite{ShortestToAllPath}, the authors study network flow optimization in the same spirit as with the $p$-resistance. Their viewpoint is based on network routing problems and provides a spectrum of routing options that make a compromise between latency and energy dissipation in selecting routes in a network. They, however, do not explicitly define a graph node distance.

The $p$-resistance distance is theoretically sound, but it lacks a closed form expression for computing all the pairwise distances of a graph. Thus, the result can only be obtained by solving a minimization 
problem for each pair of nodes separately. This currently limits the method to be applicable only for small graphs, which is why we were able to include it only in the experiments with small artificial graphs of Sections~\ref{sec:Vis}-\ref{sec:ArtificialGraphs}, but not the others.

\subsection{Other graph node distances}
In the comparisons in Section~\ref{sec:Exp}, we focus on the above presented distance families that specifically find a compromise between the SP and CT distances. However, we would like to mention other interesting distances that have been defined for graphs. In~\cite{BoP}, where the free energy distance was already presented as the potential distance, the authors also define another distance family, called the \emph{surprisal distance}. Its definition shares the same background with the free energy, but it does not generalize the SP and CT distances. However, it is shown in~\cite{BoP} to provide good results in clustering and semisupervised learning.

In~\cite{vonLuxburg-2010}, the authors define the \emph{amplified commute time distance} in order to correct the inconvenience of the commute time distance in large graphs. Of all the references above in Section~\ref{sec:OtherDists} to the work of Chebotarev, we want to highlight the \emph{walk distance}~\cite{Chebotarev-2012}, which has some interesting, quite unintuitive properties, especially when considering the peripheral nodes of a network. Finally, we mention the \emph{communicability distance}~\cite{Estrada-2012-linalg, Estrada-2012-physrev}, which measures how well a disturbance is carried over the network between two nodes when considering the network as a quantum harmonic oscillator network in a thermal bath. This model shares similarities with the RSP framework, including an inverse temperature parameter assigned to the network. Bavaud~\cite{bavaud2010euclidean} investigated the spectrum of different Euclidean distances that can be defined on weighted graphs. He also discusses the notion of \emph{focused} distances, i.e.\ distances which are zero for nodes that are adjacent to the same set of neighboring nodes. He also studies the distances by applying them in \emph{thermodynamic clustering}, which, interestingly enough, is based on minimizing a free energy functional. There also exists a vast literature on kernels on graphs, which can always be converted to matrices of squared Euclidean distances. For a thorough review on graph kernels, we advice the reader to see \cite{Fouss-2012}.

\section{Experiments}
\label{sec:Exp}
In this Section, we compare the different distance families presented in the previous Sections, namely the RSP dissimilarity, the FE distance, the $p$-resistance distance, the logarithmic forest distance and the SP-CT combination distance. First, we consider small artificial graphs and study the behavior of the different distances with different parameter values. This is done by seeing how the relation between distances of different pairs of nodes changes as the parameter value is altered. As also notified by Chebotarev~\cite{Chebotarev-2012}, the interest in comparing different distance measures does not lie in the pairwise distances themselves, but in the proportions between the pairwise distances. We then study the applicability of the distance families in a clustering experiment using networks constructed from the Newsgroups corpus of text documents. Finally, we compare the performances of the different distance families in a graph-based semisupervised learning experiment.

\subsection{Visualization}
\label{sec:Vis}
In order to show that the different distance families are appealing to use, we show a graph visualization example using an artificially generated graph. The graph is generated with the benchmark algorithm of Lancichinetti, Fortunato and Radicchi~\cite{Lancichinetti-2009}, which we refer to as the LFR algorithm. The algorithm can generate graphs that contain a community structure and obey a power law distribution both for node degrees and community sizes. The user can also set a mixing parameter which essentially means the likelihood of inter-community edges. Moreover, it is possible to include overlapping nodes that belong to several communities. The particular graph used in the visualization was generated to consist of four communities of 80 nodes, with the mixing parameter set to 0.2, the power law exponent of the degree distribution set to -2, the average degree set to 12, the maximum degree to 160 and the number of overlapping nodes, belonging to two communities, to 5.

The visualization is achieved with classical multidimensional scaling (CMDS)~\cite{ModernMDS} using the set of distances between all pairs of nodes provided by the different distance families. We show the 2-dimensional representation according to the coordinates obtained with CMDS using the SP distance, the CT distance, the RSP dissimilarity, the FE distance and the parametrized distances introduced in Sections~\ref{sec:SP-CT}-\ref{sec:pRes}. We also plot the edges of the graph with grey lines. For all the parametrized distances, we selected one intermediate parameter value, just in order to differentiate the plots from the plots obtained with the SP and CT distances. The visualizations are presented in Figure~\ref{fig:Vis} and the particular parameter values are given in the captions.

\newcommand{\LFRscalea}{.26}
\newcommand{\LFRscaleb}{.19}
\newcommand{\LFRcaptlength}{5cm}
\begin{figure}[h!]
\begin{center}
  \subfigure[\scriptsize{SP distance}]{
    \includegraphics[trim=110 200 110 200, clip=true, scale=\LFRscalea]{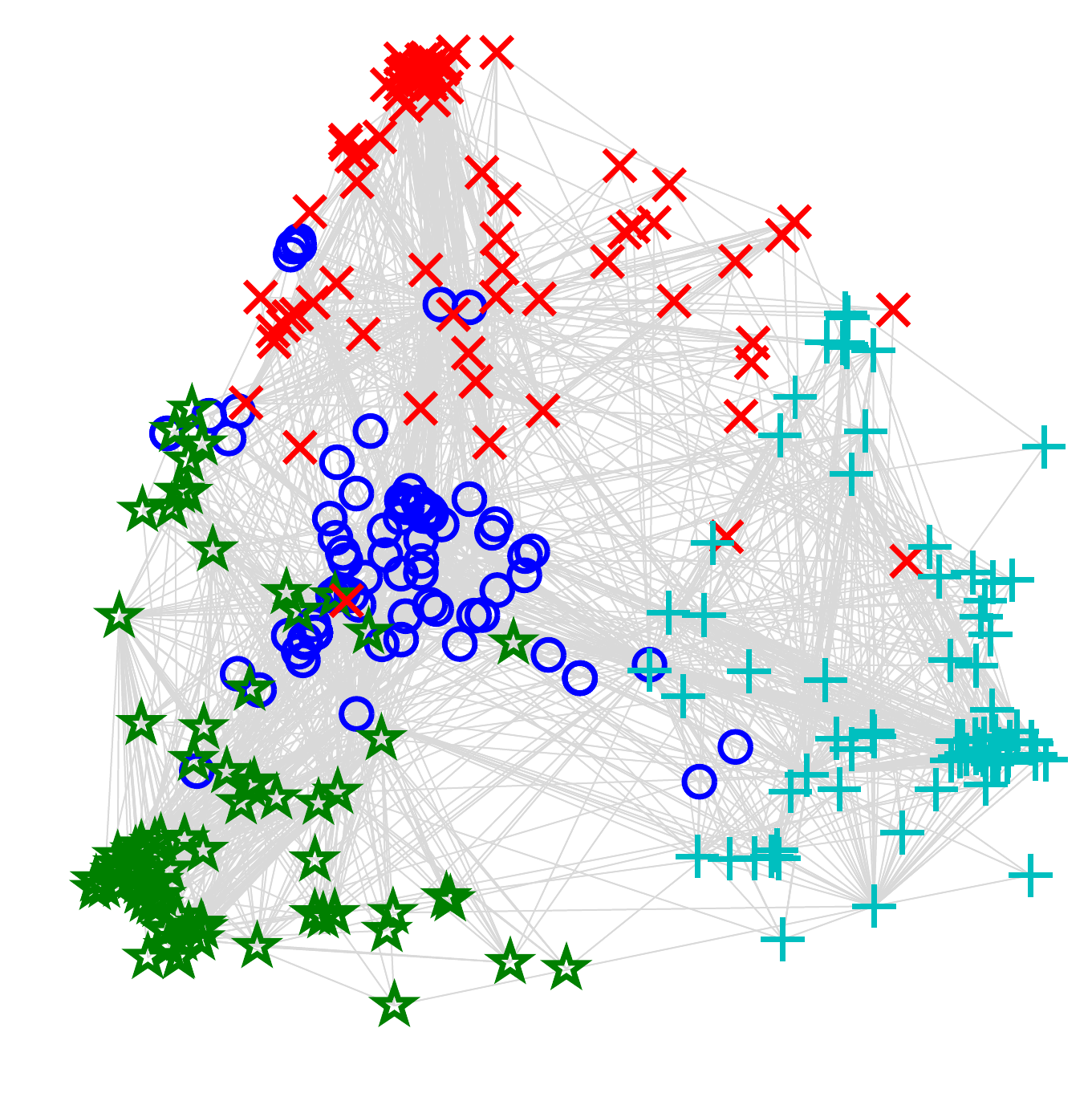}}
  \subfigure[\scriptsize{SP-CT combination distance, ${\lambda=0.1}$}]{
    \includegraphics[trim=70 200 70 200, clip=true, scale=\LFRscalea]{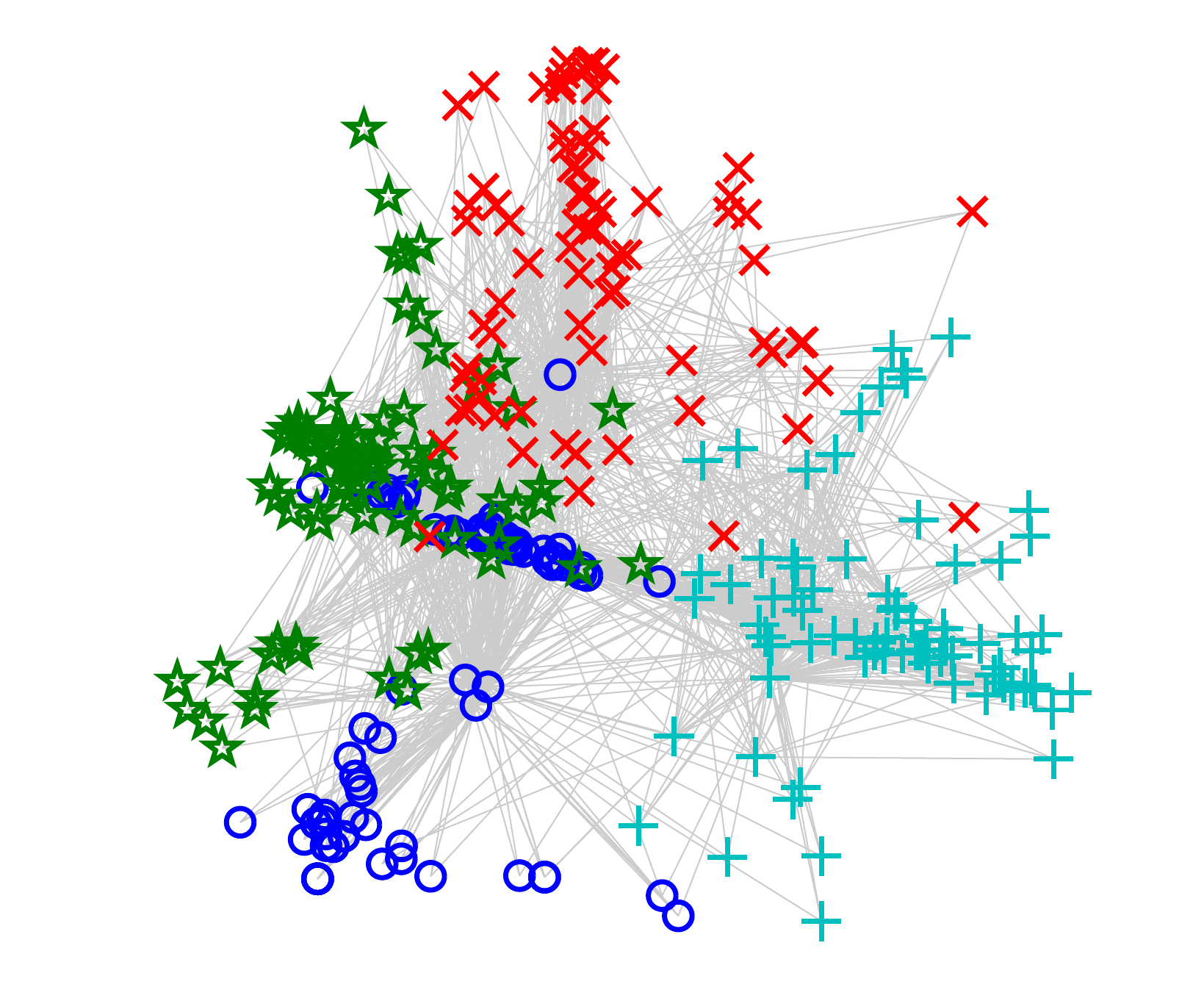}}
  \subfigure[\scriptsize{CT distance}]{
    \includegraphics[angle=90, trim=70 200 70 200, clip=true, scale=\LFRscalea]{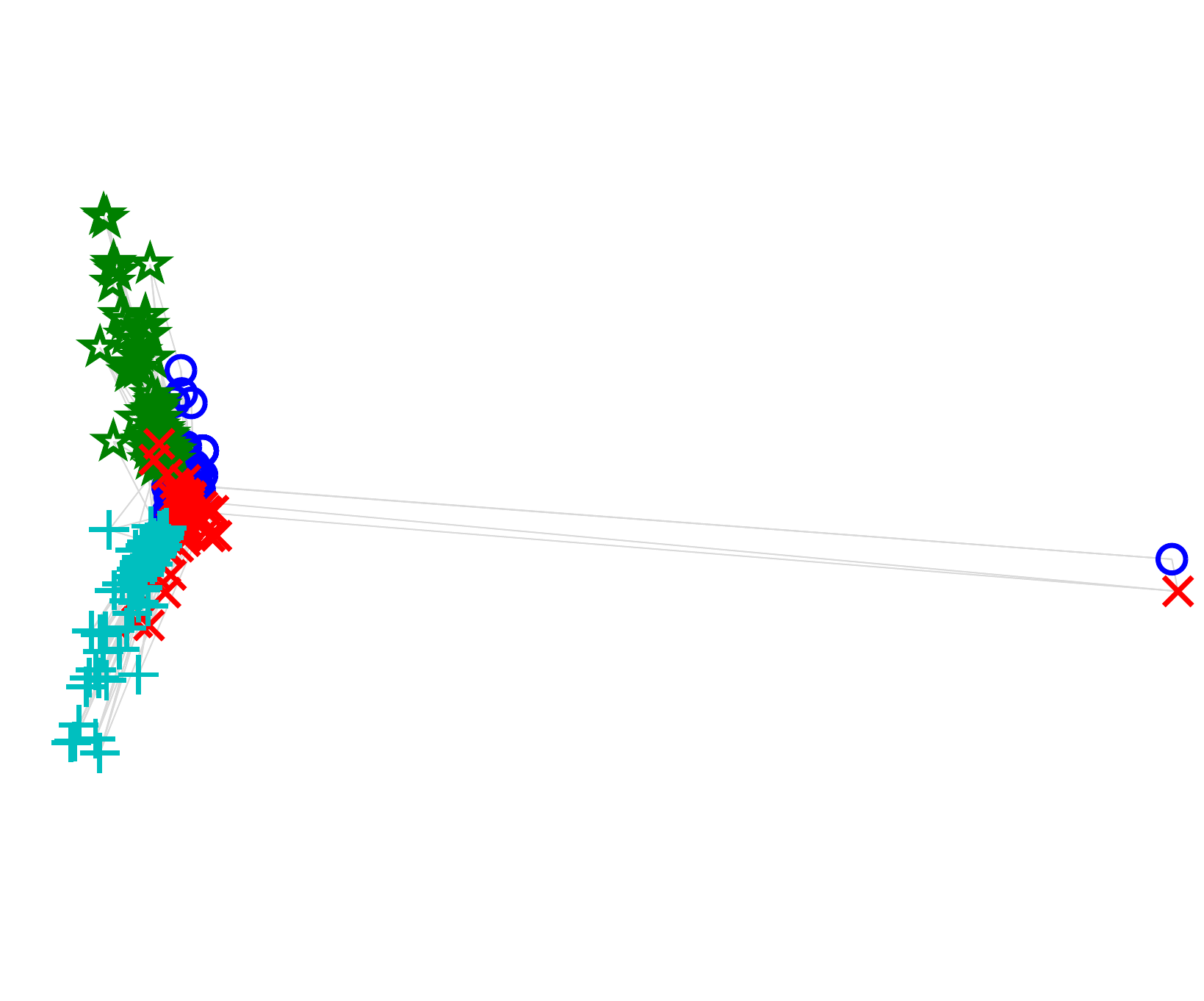}}\\
  \subfigure[\scriptsize{RSP dissimilarity, ${\beta=0.05}$}]{
    \includegraphics[trim=110 200 110 200, clip=true, scale=\LFRscaleb]{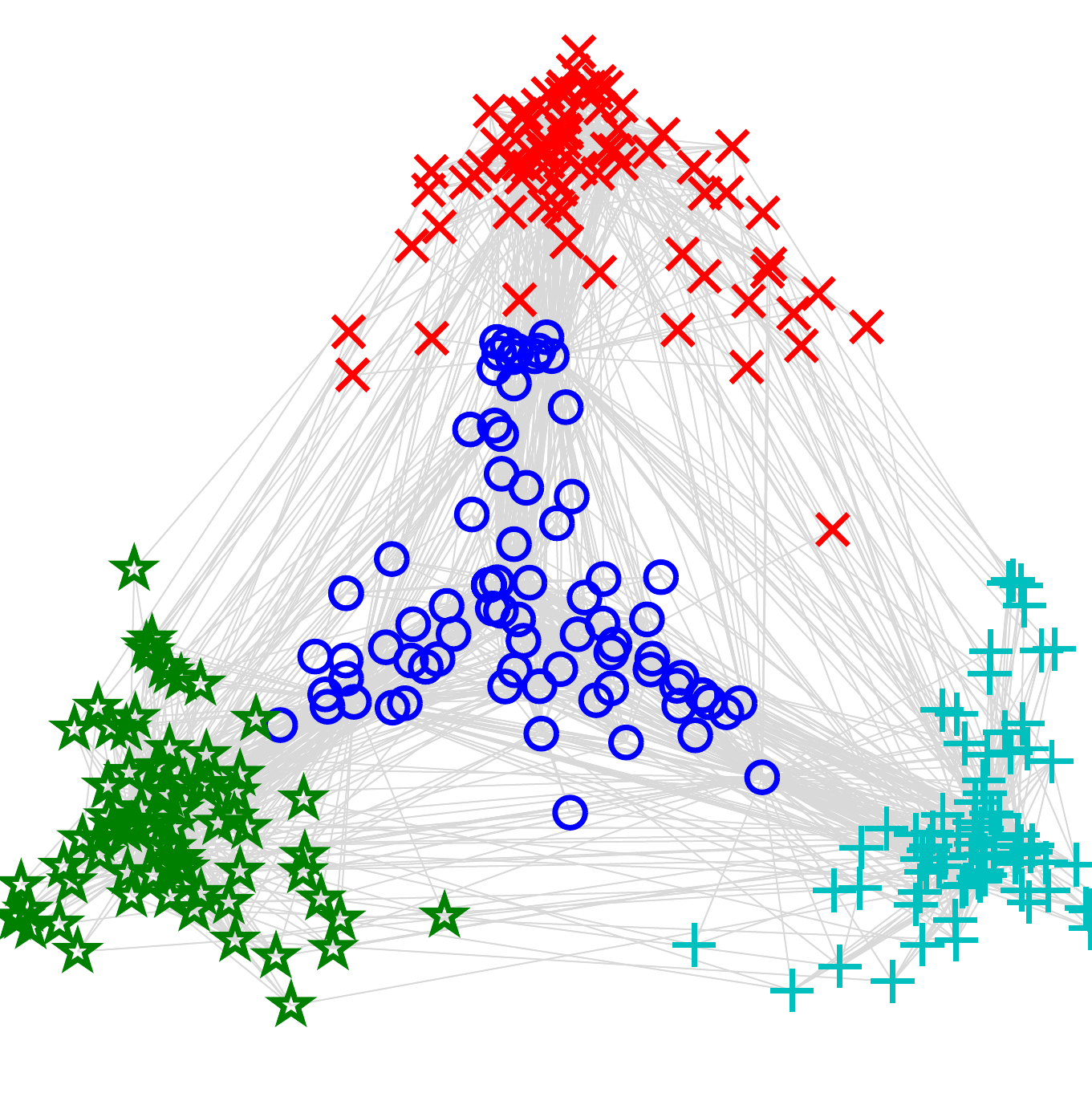}}
  \subfigure[\scriptsize{FE distance, ${\beta=0.05}$}]{
    \includegraphics[trim=110 200 110 200, clip=true, scale=\LFRscaleb]{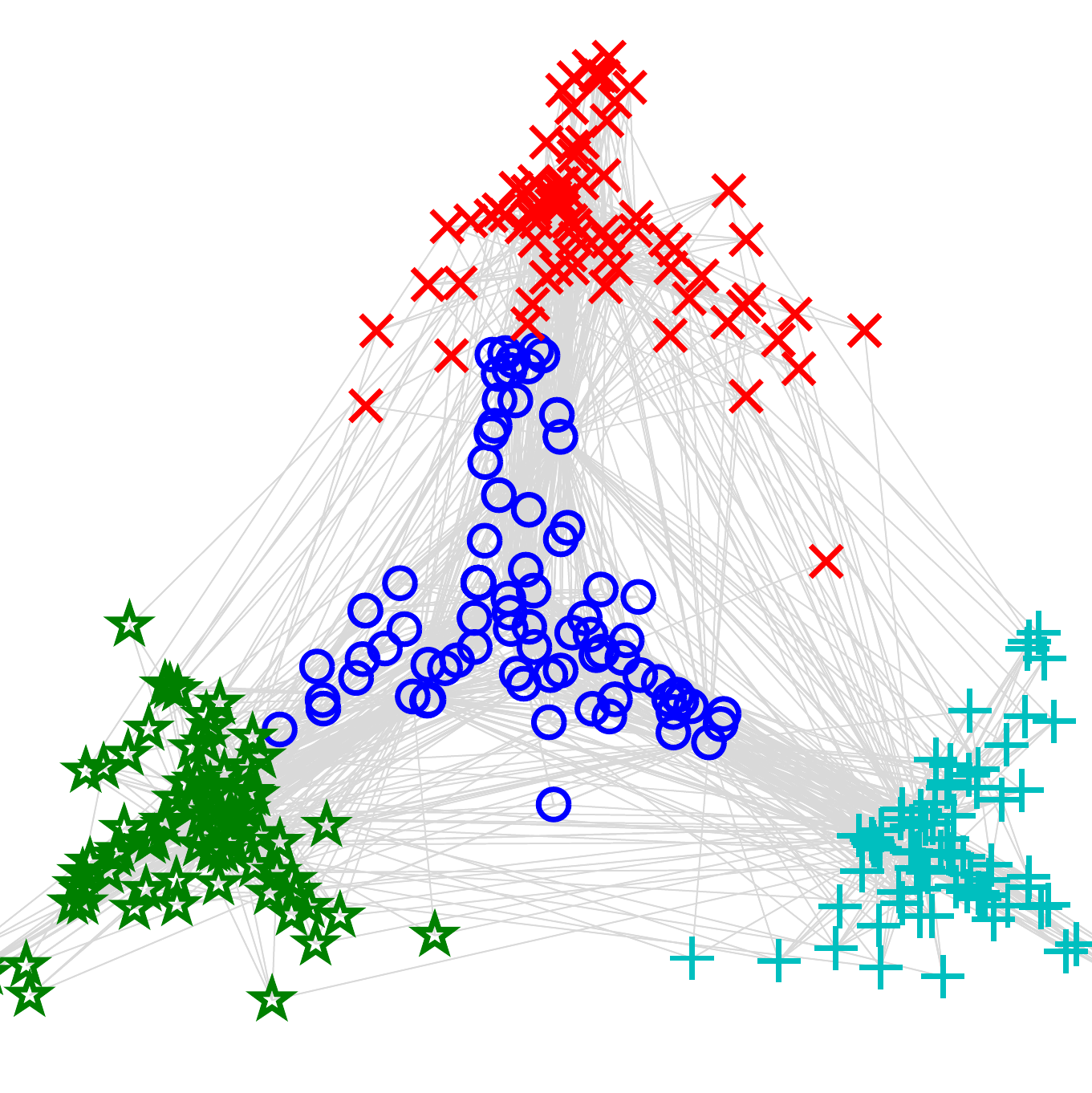}}
  \subfigure[\scriptsize{Logarithmic forest distance, ${\alpha=0.8}$}]{
    \includegraphics[trim=90 200 110 200, clip=true, scale=\LFRscaleb]{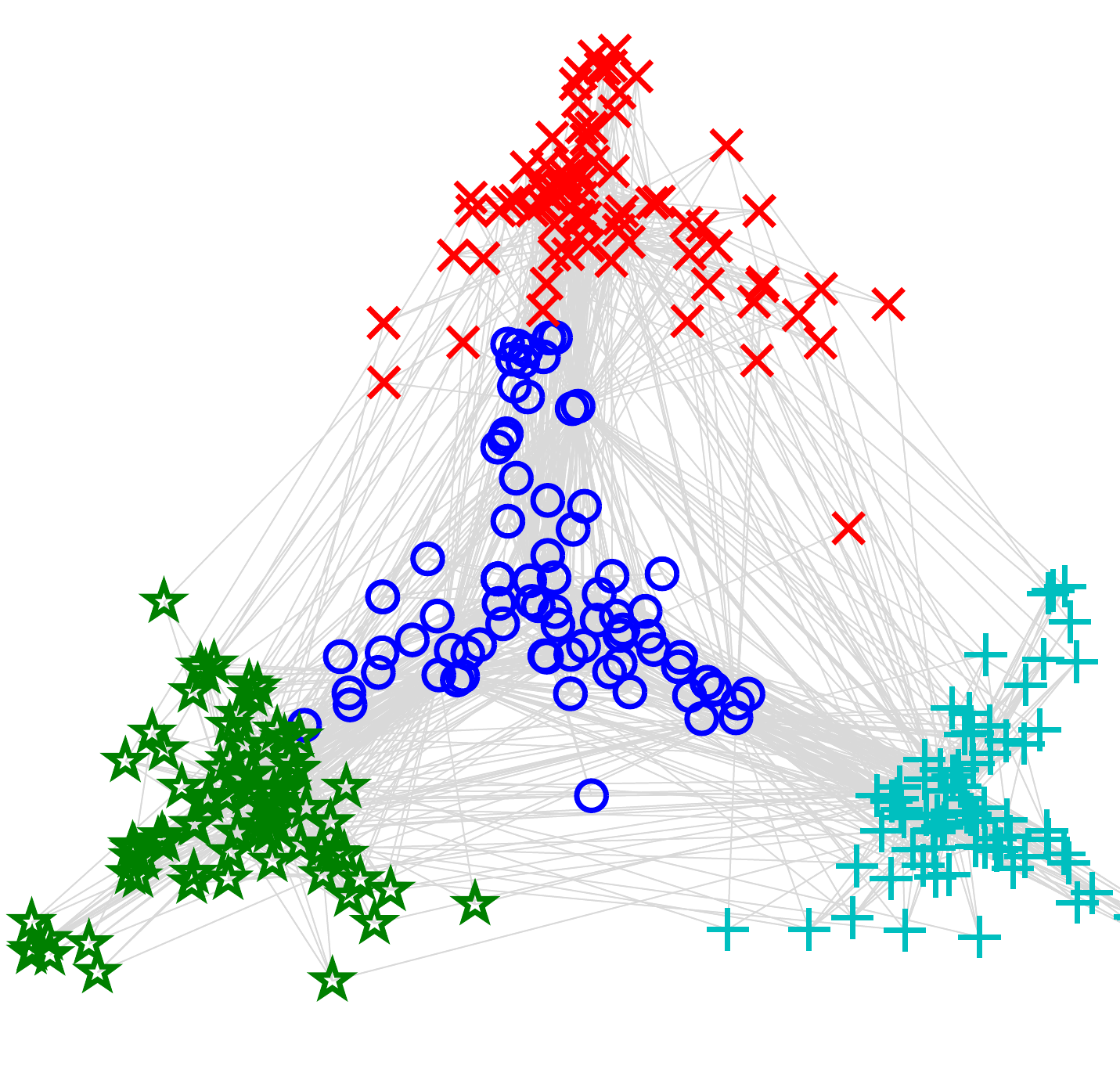}}
  \subfigure[\scriptsize{$p$-resistance distance, ${p=1.5}$}]{
    \includegraphics[trim=90 200 75 200, clip=true, scale=\LFRscaleb]{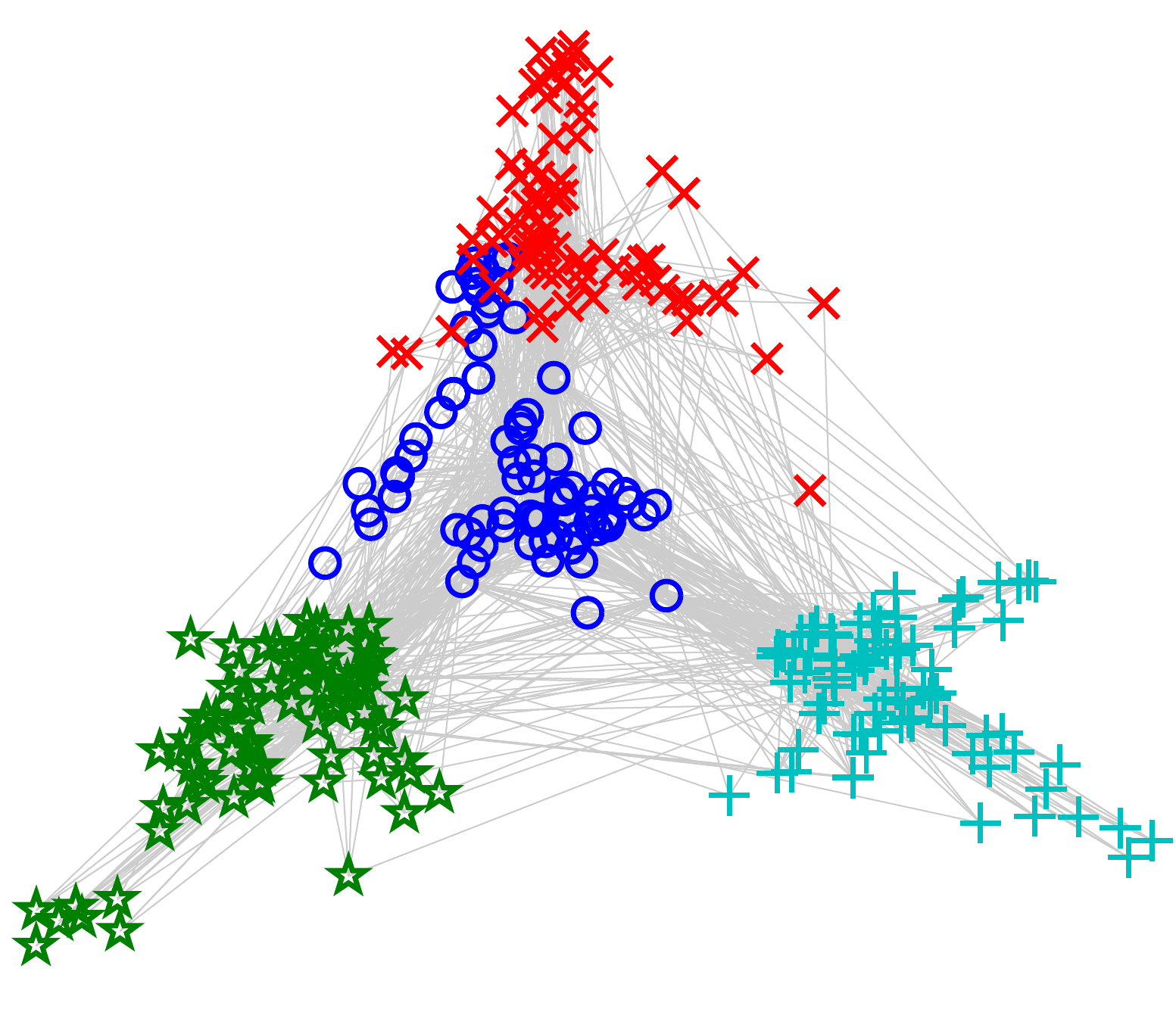}}

\caption{Visualizations with CMDS of a graph generated with the LFR algorithm consisting of four communities represented by red crosses, blue circles, green pentagrams and cyan plus-signs.}
\label{fig:Vis}
\end{center}
\end{figure}

There are two main notions that can be drawn from the plots. The first is that the parametrized distances, excluding the SP-CT combination distance, clearly manage to present the community structure better than the SP and CT distances. The plots obtained with the SP and SP-CT combination distances contain much more overlap between the communities than the other parametrized distances. The plot obtained with the CT distance also contains a lot of overlap between communities, but in addition to that, the plot is distorted by the distances of two nodes which get drawn far apart from the rest of the nodes. The degree of the two nodes is 3, which is the minimum degree of the graph. Thus, the distortion of the plot is in correspondence with the undesired dependence of the CT distance on the degrees of nodes, discussed in Section~\ref{sec:DistFams}. In fact, these nodes have, among all the nodes, the two largest sums of CT distances to all other nodes of the graph.

The second main notion is that the plots obtained with the other parametrized distances, apart from the SP-CT distance, are quite similar to each other. There is a bit more overlap between the two upper communities in the plot obtained with the $p$-resistance distance compared to the others, but this difference is very small.

%
%

\subsection{Comparisons with small graphs}
\label{sec:ArtificialGraphs}
\begin{figure}[tpb]
\centering
\subfigure[]{
    \raisebox{1.1cm}{
    \scalebox{0.9}{
    \begin{tikzpicture}[-,>=stealth,auto, node distance=4cm]
	  \tikzstyle{node} = [circle,fill=black!20,minimum size=14pt,inner sep=0pt]
	  \tikzstyle{weight} = [font=\scriptsize]
	
	  \foreach \name/\spota/\spotb/\text in {P-1/0/0/1, P-2/2/0/2, 
	                                  P-3/3.73/1/3, P-4/3.73/-1/4}
	    \node[node,draw,font=\footnotesize,xshift=0cm,yshift=0cm] (\name) at (\spota/1.3, \spotb/1.3) {$\text$};
	
	  \foreach \from/\to/\weight in {1/2/1,2/3/1,2/4/1,3/4/1}
	      \path (P-\from) edge node {} (P-\to);
	      
	\end{tikzpicture}}}
  \label{fig:ExtTri}}
\subfigure[]{
\includegraphics[scale=.6]{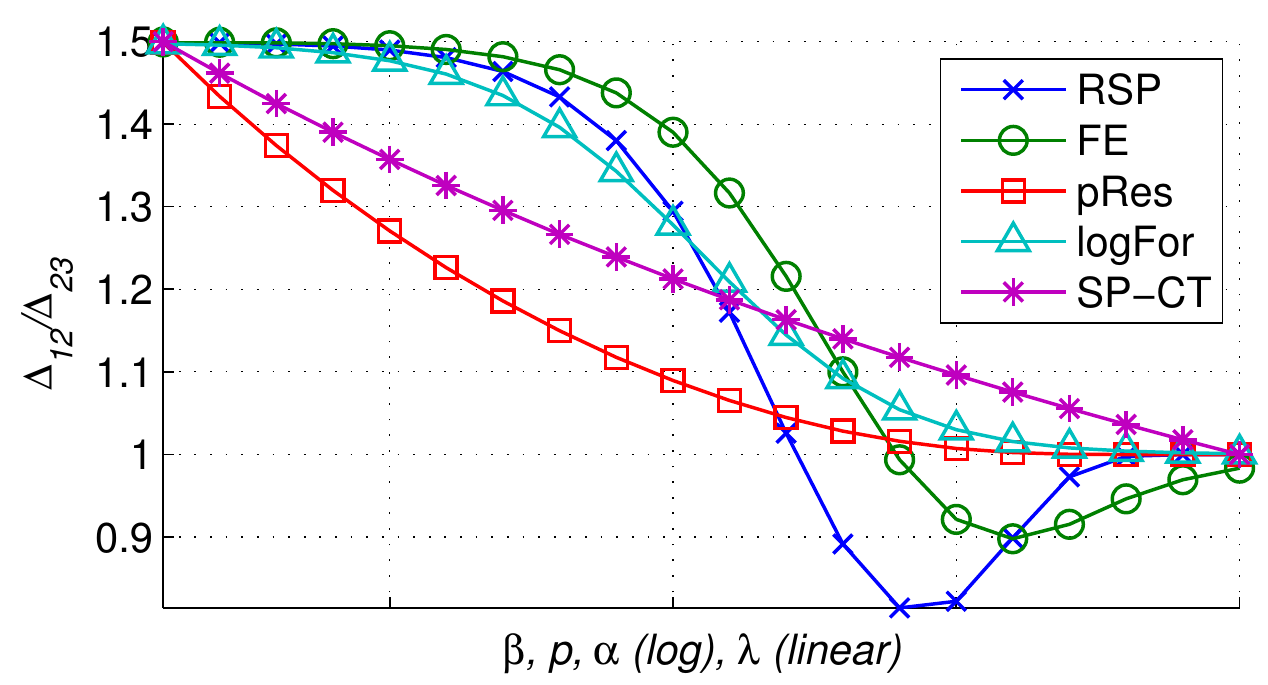}
\label{fig:ExtTriResults}}

\caption{The extended triangle graph (a) and the ratio of distances $\dist_{12}/\dist_{23}$ (b) with the RSP dissimilarity (RSP) and the FE distance (FE), both with $\beta \in [10^{-4}, 20]$, the $p$-resistance distance (pRes) with $p \in [1, 2]$ (reversed), the logarithmic forest distance (logFor) with $\alpha \in [10^{-2}, 500]$ (reversed) and the SP-CT combination distance (SP-CT) with $\lambda \in [0,1]$.}
\end{figure}
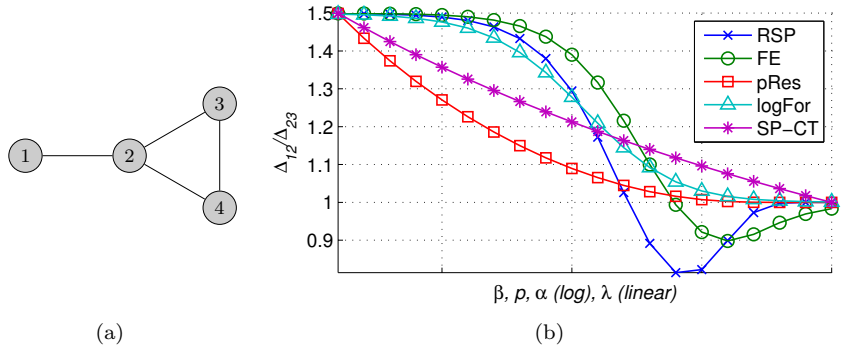

In the first example, we use the simple graph depicted in Figure~\ref{fig:ExtTri} consisting of a triangle, i.e.\ a 3-clique connected to an isolated node. We call it the extended triangle graph. We observe the proportions of distances between nodes 1 and 2 and nodes 2 and 3, i.e. the quantities $\dist_{12}/\dist_{23}$ for all the different distance families. We plot the results in Figure~\ref{fig:ExtTriResults} using 20 different parameter values for each family of distances. The parameter values are scaled in such a way that the relevant parameter range of each distance family becomes visible. In addition, the abscissa is logarithmic for all other parameters but linear for the $\lambda$ of the SP-CT combination distance.

First of all we can observe that all curves converge to unity on the right hand end of the plot. This happens as all the distances converge to the shortest path distance and thus $\dist_{12} = \dist_{23} = 1$ for all distances. On the left end of the plot, all curves approach the value $1.5$ which is the ratio of the CT distances between the nodes. In other words, for the CT distance $\dist_{12}^{\mathrm{CT}} > \dist_{23}^{\mathrm{CT}}$ holds which is caused by the fact that nodes 2 and 3 are, in a sense, better connected together (namely through node 4) than nodes 1 and 2.

The real interest in Figure~\ref{fig:ExtTriResults} lies in the transformation that takes place in the intermediate parameter values of the distance families. We can observe that the ratio $\dist_{12}/\dist_{23}$ changes monotonously with respect to the parameter value change in three cases, with the $p$-resistance, the logarithmic forest distance and obviously the SP-CT combination. In other words, these three metrics always consider the distance between nodes 2 and 3 smaller than the distance between node 2 and the isolated node 1.

However, with the FE distance and the RSP dissimilarity, the ratio behaves non-monotonously. In other words, for a range of intermediate parameter values, these functions consider the distance between the isolated node 1 and the central node 2 to be smaller than the distance between nodes 2 and 3 (and between 2 and 4). Allowing this possibility could prove useful for a distance measure in applications. For example, in a social network, a relationship with an isolated person can in some situations and contexts be considered stronger than the relationship with a member of a group.


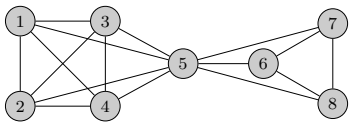
\begin{figure}[]
\centering
\scalebox{0.8}{
    \raisebox{0cm}{\begin{tikzpicture}[-,>=stealth,auto, node distance=4cm]
	  \tikzstyle{node} = [circle,fill=black!20,minimum size=14pt,inner sep=0pt]
	  \tikzstyle{weight} = [font=\scriptsize]
	
	  \foreach \name/\angle/\text in {P-1/135/1, P-2/-135/2,
	  																 P-3/45/3,  P-4/-45/4}
	    \node[node,draw,font=\footnotesize,xshift=0cm,yshift=0cm] (\name) at (\angle:1cm) {$\text$};

	  \foreach \from/\to/\weight in {1/2/1,1/3/1,1/4/1,2/3/1,2/4/1,3/4/1}
	      \path (P-\from) edge node {} (P-\to);
	  
	  \foreach \name/\spotx/\spoty/\text in {P-5/0/0/5, P-6/2/0/6, 
	                                  P-7/3.73/1/7, P-8/3.73/-1/8}
	    \node[node,draw,font=\footnotesize,xshift=2cm,yshift=0cm] (\name) at (\spotx*2/3, \spoty*2/3) {$\text$};
	
	  \foreach \from/\to/\weight in {6/7/1,6/8/1,7/8/1}
	      \path (P-\from) edge node {} (P-\to);
	      
	  \foreach \from/\to/\weight in {1/5/1,2/5/1,3/5/1,4/5/1,6/5/1,7/5/1,8/5/1}
	      \path (P-\from) edge node {} (P-\to);
	      
	\end{tikzpicture}}}
\caption{A graph with a 4-clique and a 3-clique and a hub node between them.}
\label{fig:Unbal}
\end{figure}

The phase transition that occurs with the FE distance and the RSP dissimilarity in our small example case can have implications in more practical situations as well. Obviously, it can affect nearest neighbor related methods but also clustering applications. Consider, for example, a larger scale situation, as the graph depicted in Figure~\ref{fig:Unbal}. This graph consists of two cliques of sizes 4 and 3 which are connected through a hub node (node 5) that shares edges with all the other nodes of the graph. Consider then a clustering of the graph nodes into two clusters. The nodes in the two cliques obviously should belong to their own clusters. But which cluster should the hub node 5 be assigned to? This is generally a question of context and taste. In some cases there might be a preference for classifying the hub node to the smaller cluster, whereas in others it should be considered part of the larger cluster. One option would also be to put the hub node into its own cluster. However, here we are interested in cases where the number of clusters is fixed and a decision on the cluster assignment of the hub node has to be made.

In this specific case, the $p$-resistance distance, the logarithmic forest distance and the SP-CT combination distance always consider node 5 closer to the larger clique than the small one. Thus, for example, when performing a $k$-means based clustering with $k=2$, using the mentioned distances will always result in assigning the hub node 5 into the larger cluster. However, the other three distances are more flexible. Namely, thanks to the phase transition seen in Figure~\ref{fig:ExtTriResults}, performing $k$-means with these distances can result in two different partitions depending on the parameter value. Worth pointing out is that since the shortest path distance between node 5 and all other nodes is 1, a $k$-means clustering can result in either of the two interesting partitions, because with both partitions the global minimum within-cluster inertia is achieved. 


These examples illustrate that there are subtle differences between the generalized distance families. These differences may be useful for deciding which distance measure should be used in which case and they might give insight about how to select the parameter value of a generalized distance depending on the nature and properties of the data. In Sections~\ref{sec:Clus1} and~\ref{sec:NN}, we test the different distance measures in clustering and semisupervised learning in order to observe the capabilities of the different distance families in detecting desired clusters in data. In the future we will extend this investigation to other problems such as link prediction~\cite{Liben-Nowell-2007, Lu-2011}.

\subsection{Graph node clustering}
\label{sec:Clus1}

In order to see whether the distance families in fact achieve to extract a meaningful representation of a graph, we use them in a graph node clustering task and evaluate the quality of the partitions found. For the clustering, we employ the kernel $k$-means algorithm introduced in~\cite{SigCT}. It is based on an iteration procedure similar to $k$-means that searches for prototype vectors, but in a \emph{sample space} and by using a kernel containing similarities between data points, instead of using explicit data vectors in a feature space. This is convenient in our case, because we only possess information of distances between data points instead of a vector representation. The dimension of the sample space is $n$, i.e.\ the number of data points, and each data point is represented in the sample space as the corresponding column vector of the kernel. Inner products between the data and prototype vectors in the sample space correspond to distances in an \emph{embedding space} by application of the \emph{kernel trick}~\cite{Shawe-Taylor-2004}. The goal of the algorithm is to find prototype vectors in the sample space that minimize the within-cluster inertia in the embedding space, i.e.\ the sum of the squared distances of each data point to its corresponding prototype.

In order to use the kernel $k$-means algorithm we need to convert each matrix of distances into a matrix of similarities. We transform the matrices of distances $\mathbf{\Delta}$ into similarity matrices $\mathbf{K}$ in a classical way~\cite{ModernMDS,schoenberg1938metric,gower1986metric} as
\begin{equation}
\label{eq:MinusDoubleCenter}
\mathbf{K} = -\dfrac{1}{2}\mathbf{H \Delta} \mathbf{H},
\end{equation}
where $\mathbf{H} = \mathbf{I} - \mathbf{ee}^{\mathsf{T}}/n$ is the centering matrix. When the matrix $\mathbf{\Delta}$ contains squared Euclidean distances, matrix $\mathbf{K}$ will contain inner products of centered vectors in the same Euclidean space. We use the distances without raising them to the square, because the commute time distance already is the square of a Euclidean distance, called simply the Euclidean commute time distance~\cite{MarcoPCA}. However, for intermediate parameter values, the generalized distances are not necessarily Euclidean (or squared Euclidean) distances and thus the corresponding similarity matrices are not necessarily kernels in the traditional sense as they might not be positive definite. The positive definiteness could be ensured by forcing the negative eigenvalues of the similarity matrix to zeros. However, we have noticed in experiments that this does not affect much the results nor the convergence of the kernel $k$-means. 

In addition to the similarity matrices derived from the generalized graph distance matrices through~(\ref{eq:MinusDoubleCenter}), we also use the \emph{sigmoid commute time kernel} proposed by Yen et al.~\cite{SigCT}. They construct the kernel by taking a sigmoid transformation of the elements of the commute time kernel which can be computed as the Moore-Penrose pseudoinverse $\mathbf{L}^{+}$ of the graph Laplacian. Thus, the similarities given by this method are $(\mathbf{K}^{\sigma \mathrm{CT}})_{st} = 1/(1 + \exp(-a\, l_{st}^{+} / \sigma)).
$
The parameter $a$ controls the smoothing of the similarity values caused by the sigmoid transformation and $\sigma$ is the standard deviation of the elements $l_{ij}^{+}$. The sigmoid commute time kernel has been shown to perform well in many machine learning tasks, especially in the kernel $k$-means method used in this paper. We consider it as a baseline for the clustering performance comparisons. 

We use a collection of text document networks extracted from the 20 Newsgroups data set\footnote{\url{http://qwone.com/~jason/20Newsgroups/}}. A more detailed description of the collection that we use can be found 
in~\cite{SigCT}. In short, our collection consists of ten different weighted undirected networks, where the nodes represent text documents and edges and their affinities are formed according to the co-occurrence of words within the documents. The affinities $a_{ij}$ have been converted to costs $c_{ij}$ as $c_{ij} = 1/a_{ij}$. Each network has been constructed by combining subsets of 200 documents from one topic. The networks consist of either two, three or five of such subsets of documents resulting in networks of 400, 600 and 1000 nodes. The goal of the clustering is then to detect the division of each network according to the topics. Unfortunately, we could not obtain results in this experiment with the $p$-resistance because of its high computational cost discussed already in Section~\ref{sec:pRes} and the sizes of the networks in the experiment.

We compare the partitions found by the clusterings with the classification according to the topics by computing the \emph{normalized mutual information} (NMI)~\cite{Strehl02clusterensembles}
between a clustering partition $X$ and the topic classification $Y$ as
\[
NMI(X,Y) = \dfrac{I(X,Y)}{\sqrt{H(X)H(Y)}},
\]
where $I(X,Y)$ is the mutual information of partitions $X$ and $Y$ and $H$ is entropy. In order to avoid the expensive running of the experiments every time for a wide range of parameter values, we perform a simple parameter tuning. We assume that the tuning results generalize from one data set of a particular kind (here, a text document collection) to another. We separate one of the ten networks in our collection as a tuning data network. We perform the kernel $k$-means clustering for this network with 20 different random initializations of the prototype vectors. Out of these 20 runs, we observe the NMI score of the clustering that achieves the smallest within-cluster inertia. This process is repeated another 20 times for a set of parameter values distributed either logarithmically (or linearly, in the case of the SP-CT-distance) on a given range of values. The parameter value producing the largest mean NMI score is chosen as the tuned value. The same procedure is done for all the distance measures as well as the sigmoid CT kernel. 

The ranges of parameter values and the optimal values for each method are reported in Table~\ref{tab:ParVals}. Note that with the SP-CT combination distance, the optimal parameter value is $\lambda = 1$. This means that already for the 600 node tuning network the best clustering results with the SP-CT distance are obtained when focusing only on the SP distance and neglecting the CT distance.

\begin{table}[t]
\begin{center}
\scalebox{0.7}{%
\begin{tabular}{l|l|l|c}
Distance & Similarity matrix & Parameter range & Optimal value \\
 \hline
 RSP dissimilarity           & $\mathbf{K}_{\mathrm{RSP}}$    & $[10^{-4}, 20]$ & $\beta = 0.02$ \\
 FE distance        & $\mathbf{K}_{\mathrm{FE}}$     & $[10^{-4}, 100]$ & $\beta = 0.07$ \\
 Logarithmic forest distance & $\mathbf{K}_{\mathrm{LogF}}$   & $[10^{-2}, 500]$ & $\alpha = 0.95$ \\
 SP-CT combination           & $\mathbf{K}_{\mathrm{SP-CT}}$  & $[0, 1]$ & $\lambda = 1$ \\
 Sigmoid commute time        & $\mathbf{K}_{\mathrm{SigCT}}$  & $[10^{-2}, 10^{3}]$ & $a = 26$ \\
\end{tabular}}
\caption{The notation of the similarity matrices and the optimal parameter values obtained on the tuning data set.}
\label{tab:ParVals}
\end{center}
\end{table}

We then use the tuned parameter values to perform the clustering on the nine remaining networks. As in the tuning phase, we again perform the clustering with 20 different initializations and choose the clustering that has the smallest within-cluster inertia. This is again done another 20 times and the mean and standard deviations of the NMI scores of these 20 best clusterings are collected. The results for each of the nine data sets are reported in Table~\ref{tab:NewsRes}. For each data set, we performed one-sided $t$-tests with significance level 0.05 to determine whether a mean NMI score with one method is significantly higher than with another, which is indicated in boldface.

From the results we see that the highest NMI scores are generally obtained with the similarity matrices based on the FE distance, the RSP dissimilarity and the sigmoid commute time kernel. We can see that the RSP dissimilarity and the FE distance provide scores very close to each other. The clusterings obtained based on the logarithmic forest distances do not correspond that well to the topic labeling except with the network G-2cl-B, for which all the distances, excluding the SP distance, give quite similar results. In general, all the distance families provide NMI scores that are quite close to each other, with the exception that with some networks, the score provided by the SP distance is clearly smaller than the others. Thus, the experiment indicates that the parametrized distances, when used with intermediate parameter values, do manage to capture more meaningful global information of the graph structure than the SP and CT distance do.

We also took a more detailed look at the individual cluster assignments of nodes to compare the partitions obtained with the different similarity matrices. We can only describe the results because a thorough presentation would require too much space. First of all, there are nodes in almost all of the networks whose cluster assignment never corresponds to the topical classification with any of the similarity matrices. This only indicates that the data is noisy and that the topical classification cannot be perfectly inferred from the network structure. Also, all the similarity matrices produce a clustering of the network G-5cl-2 where the first two classes are clustered mostly together and one of the topical classes is divided into two smaller clusters. This indicates a hierarchical structure of the classes and it seems to be interpreted by the different similarity matrices in different ways.

The similarity of the NMI scores between the RSP dissimilarity and the FE distance is explained by looking at the respective cluster assignments, which indeed are almost exactly similar with all the networks. However, the difference between the partitions based on these two distances and the others is evident from the individual assignments. Interestingly, especially with the larger networks, the clusterings based on the logarithmic forest distance and the sigmoid commute time kernel share some similarities. These similarities are most evident in the clusterings of the network G-5cl-2, mentioned above. Also, with this network, the clustering produced by the SP distance is in fact quite close to the ones produced by the RSP dissimilarity and the FE distance. All in all, the individual cluster assignments indicate many small differences in the clusterings with different similarity matrices. However, in order to get a stronger intuition of the reasons causing the differences, we would need an even more thorough investigation which we leave for future work.


\begin{table}
\begin{center}
\scalebox{0.7}{%
\begin{tabular}{l c c c c c c c c c}
\hline
NMI & $\mathbf{K}_\mathrm{RSP}$ & $\mathbf{K}_\mathrm{FE}$ & $\mathbf{K}_\mathrm{LogF}$ & $\mathbf{K}_\mathrm{SP-CT}$ & $\mathbf{K}_\mathrm{SigCT}$\\
Datasets & & & & &\\
\hline
G-2cl-1 & {\bf 84.5}  $\pm$  0.00 &  80.7  $\pm$ 1.09 &  83.1  $\pm$ 1.47  &  65.2  $\pm$  0.59 &  81.6 $\pm$ 0.00\\
G-2cl-2 & {\bf 58.7}  $\pm$  0.38 &  {\bf 58.7}  $\pm$ 1.74 &  {\bf 58.8}  $\pm$ 1.94  &  51.2  $\pm$  0.46 &  56.8 $\pm$ 2.18\\
G-2cl-3 & 81.0  $\pm$  0.00 &  81.1  $\pm$ 0.00 &  75.0  $\pm$ 1.13  &  {\bf 85.9}  $\pm$  0.00 &  79.6 $\pm$ 0.00\\       
\\                                                                                                           
G-3cl-1 & 76.6  $\pm$  0.00 &  76.2  $\pm$ 0.00 &  75.4  $\pm$ 0.72  &  74.2  $\pm$  0.28 &  {\bf 77.3} $\pm$ 0.00\\
G-3cl-2 & 77.0  $\pm$  0.00 &  {\bf 78.3}  $\pm$ 0.83 &  75.5  $\pm$ 1.42  &  62.6  $\pm$  0.51 &  73.0 $\pm$ 0.00\\
G-3cl-3 & 76.5  $\pm$  0.28 &  {\bf 77.0}  $\pm$ 0.50 &  74.4  $\pm$ 1.57  &  71.5  $\pm$  0.50 &  75.9 $\pm$ 0.43\\     
\\                                                                                                           
G-5cl-1 & {\bf 69.6}  $\pm$  0.15 &  69.0  $\pm$ 0.66 &  60.4  $\pm$ 3.43  &  68.1  $\pm$  0.43 &  66.8 $\pm$ 0.16\\
G-5cl-2 & 64.0  $\pm$  0.42 &  {\bf 64.6}  $\pm$ 0.34 &  58.7  $\pm$ 3.49  &  59.6  $\pm$  0.59 &  60.4 $\pm$ 1.36\\
G-5cl-3 & {\bf 61.2}  $\pm$  0.71 &  {\bf 61.6}  $\pm$ 0.87 &  57.3  $\pm$ 2.77  &  47.8  $\pm$  0.92 &  57.3 $\pm$ 0.46\\
\end{tabular}}
\caption{Clustering performances (Normalized Mutual Information, multiplied by 100) for each similarity matrix on the nine Newsgroup subsets} 
\label{tab:NewsRes}
\end{center}
\end{table}


\subsection{Semisupervised 1-nearest-neighbor classification}
\label{sec:NN}
The kernel $k$-means algorithm used in the clustering experiment in the previous section considers the distances of the graph globally as it aims to minimize the within-cluster inertia. We also wanted to employ the different distance families in a semisupervised learning algorithm based on nearest neighbors, i.e.\ only on local distances. In order to observe differences between the distance families, we use the propagating 1-nearest-neighbor algorithm~\cite{Zhu-2009}, where, iteratively, the unlabeled node that is the closest to any labeled node gets assigned the label of the labeled node. This is obviously not a state-of-the-art method, but it provides a comparison of the distance families when only the most local distances are being applied.

We conduct the semisupervised learning experiments for the same Newsgroups datasets as in the clustering experiments in Section~\ref{sec:Clus1} but also for another collection of networks, the WebKB data set~\cite{Macskassy-2007}. It is a collection of four co-citation networks collected from the websites of four American universities. The nodes in the networks are webpages and the edges of the network represent co-citations, meaning that two pages are connected by an edge, if they contain a link to the same target page. We perform the propagating 1-NN algorithm using the RSP dissimilarity, the FE distance, the logarithmic forest distance and the SP-CT combination distance. Again, as was the case with the the clustering experiments in Section~\ref{sec:Clus1}, we cannot use the $p$-resistance in the comparison, because of its slow computation.

We perform the learning task with five different labeling rates, 10\%, 30\%, 50\%, 70\% and 90\%. For each labeling rate, we average the score over 20 different randomized label deletions and for each label deletion the score is given by the mean classification rate according to a 10-fold cross validation. Within each fold of the cross-validation, we run an inner 10-fold cross validation to define the optimal parameter values for each distance family. Thus, the parameter tuning is performed in a very different manner compared to the clustering experiments, which makes more sense in the semisupervised setting.

\begin{figure}[t!]
\begin{center}
\includegraphics[scale=.55]{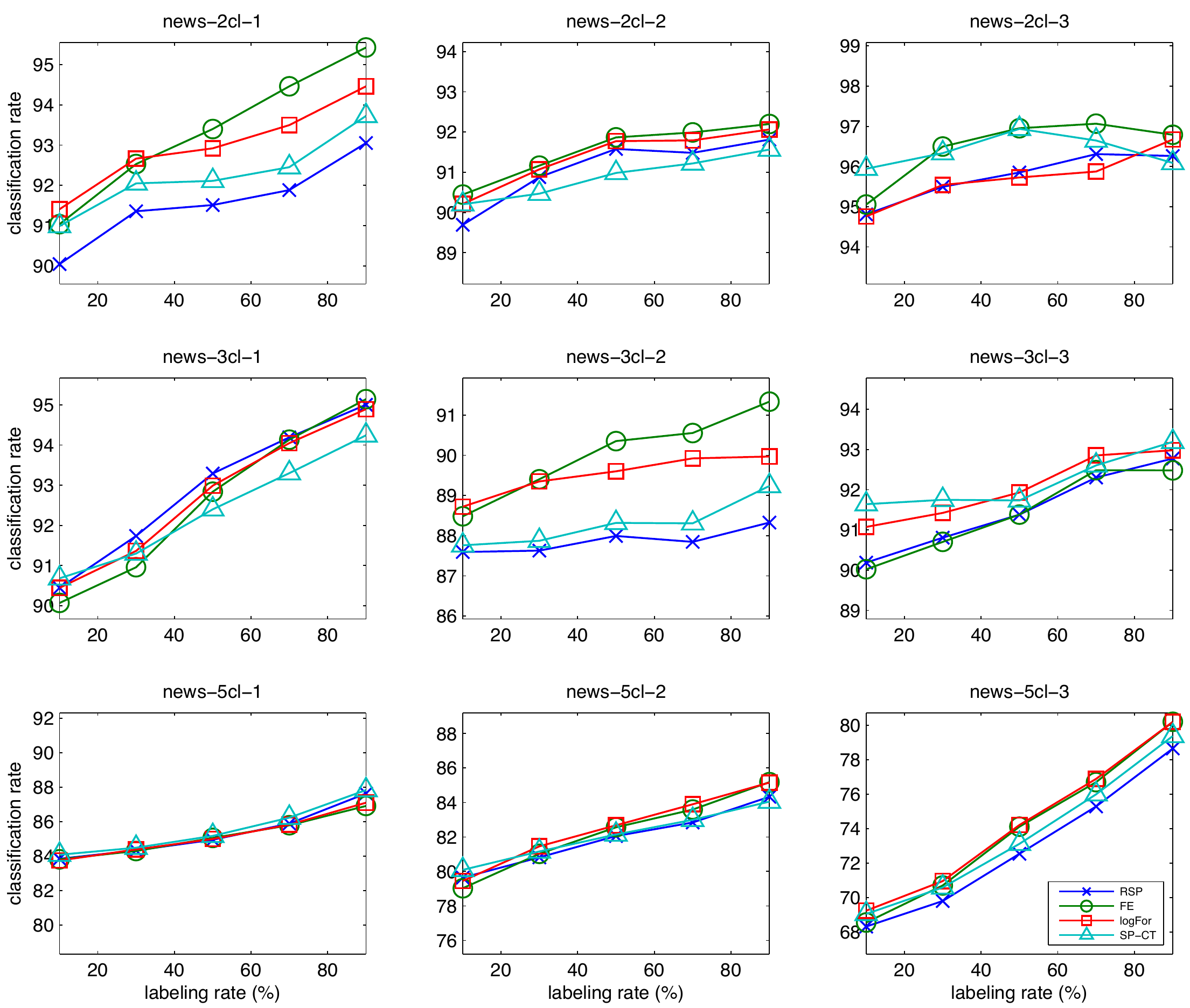}
\includegraphics[scale=.55]{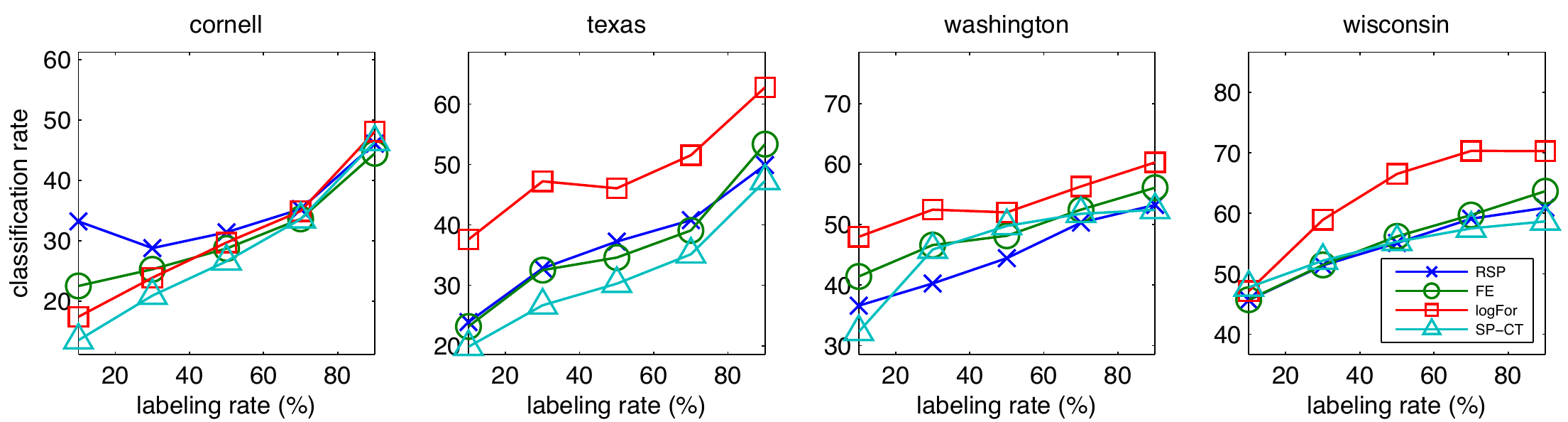}
\caption{The results of the propagating 1-NN algorithm with the nine Newsgroups graphs (top) and the four WebKB co-citation graphs (below)}
\label{fig:NNNewsRes}
\end{center}
\end{figure}

Figure~\ref{fig:NNNewsRes} shows the results of the propagating 1-NN algorithm with the nine Newsgroups data sets and four WebKB data sets. With some of the graphs, the results obtained with different distance families seem very similar to each other. However, where there is a noticeable difference, in fact the results obtained using the logarithmic forest distance seem quite good. On the other hand, the results obtained with the RSP dissimilarity are in many cases worse than the others. In other words, the results seem a bit contrary to the results obtained in the clustering experiments.

We perform a ranking of the distance families using Copeland's voting method~\cite{CopelandsMethod}, by computing the times that one distance family provides a significantly better result than another one with a 0.05 significance level. Copeland's method simply gives a score of $+1$ to a distance family that achieves a significantly superior score than another one on a given data set, and correspondingly a score of $-1$ to the other one. If there is no significant difference between two methods, they both are assigned a score 0. The ranking of the methods is then computed by summing the scores over all pairwise comparisons of methods and over all data sets.

We compute the ranking separately for each labeling rate by summing the scores over all data sets. Thus, for one labeling rate, the maximum score that a distance family can get is 39 (13 data sets and comparisons with 3 other distances). The ranks and scores obtained with Copeland's method are shown in Table~\ref{tab:NNCopeland}. They confirm the observation that the logaritmic forest distance provides good results whereas using the RSP dissimilarity results in more misclassifications. The FE distance seems to perform quite well also, being ranked second, and the SP-CT combination distance is ranked third.

The reason why the RSP dissimilarity performs so poorly in this task is a bit surprising, at least considering its suitability for clustering which was showed earlier. This result may have something to do with the fact that in some cases the RSP dissimilarity does not satisfy the triangle inequality. The triangle inequality ensures an intuitive relationship between local and global distances, literally that a sum of local distances should be larger than the corresponding global distance. Another way to interpret this is that when the inequality is not satisfied, the local distances can be smaller than a metric global topology would allow. The 1-NN algorithm is based on purely local properties of the space, whereas the kernel $k$-means algorithm considers the distances more globally, by depending on the within-cluster inertia. However, this is only a suggestion for explaining the results and we have not been able to verify it theoretically, nor by demonstration.


\begin{table}
\begin{center}
\scalebox{0.65}{%
\begin{tabular}{|l|c c|c c|c c|c c|c c||c c|}
\hline
Labeling rate & & \hspace{-1 cm} 10\% & & \hspace{-1 cm} 30\% & & \hspace{-1 cm} 50 \% & & \hspace{-1 cm} 70 \% & & \hspace{-1 cm} 90 \% & & \hspace{-1 cm} Overall \\
\hline
Sim.\ matrix & Rank & Score & Rank & Score & Rank & Score & Rank & Score & Rank & Score & Rank & Score \\
$\mathbf{K}_\mathrm{RSP}$     &  4  & -14   &  4  & -16   &  4  & -16   &  4  & -17   &  3  & -14 & 4 & -77 \\
$\mathbf{K}_\mathrm{FE}$      &  3  &  -5   &  2  &   2   &  2  &   9   &  2  &  12   &  2  &  11 & 2 &  29 \\
$\mathbf{K}_\mathrm{LogF}$    &  1  &  11   &  1  &  17   &  1  &  18   &  1  &  20   &  1  &  21 & 1 &  87 \\
$\mathbf{K}_\mathrm{SP-CT}$   &  2  &   8   &  3  &  -3   &  3  & -11   &  3  & -15   &  4  & -18 & 3 & -39 \\
\hline
\end{tabular}}
\caption{The ranking of the distance families according to Copeland's method based on the results in the propagating 1-NN learning task. The results are presented for each labeling rate separately across all data sets ($10\%, \ldots, 90\%$), and for all labeling rates together (Overall).}\label{tab:NNCopeland}
\end{center}
\end{table}

\section{Conclusion}
\label{sec:Concl}
In this article, we concentrated on graph node distances that generalize the SP and CT distances. We first presented the setting of the paper where we concentrate on graphs that have a structure based on edge weights and another based on edge costs. Within this setting, we also proved that the commute cost distance is equal to the commute time distance up to a constant factor. This small result is interesting because it means that changing a cost of one edge of the graph has the same effect for all the pairwise commute cost distances in the graph, remembering that the transition probabilities defining the corresponding Markov chain are independent of the costs.

We then developed the theory behind one parametrized distance family, the RSP dissimilarity, by providing a new closed form algorithm for computing all pairwise dissimilarities of a graph. In addition, we derived the FE distance based on the Helmholtz free energy. Although we show that the FE distance coincides with the potential distance, proposed earlier, our new derivation provides a solid theoretical background to the distance. The derivation also reveals a closer resemblance of the FE distance with the RSP dissimilarity. The significant difference between the two is that only the FE distance is a metric. The FE distance has other nice features as well, including the graph-geodetic property and a closed form solution for computing all pairwise distances.

The other focus of the article was to compare different generalized graph node distances. We gave simple examples of subtle differences between some of the distance families using small artificial graphs. We then employed the distances on different network data analysis tasks in order to compare them and to evaluate their applicability. Graph visualizations obtained with the generalized distances indicated that they manage to respect the community structure of the graph much better than the traditional distances. Clusterings based on the parametrized distances with the tuned parameter values corresponded more to the inherent topic-induced classification of the graph nodes than the clustering found based on the SP and CT distances. We also performed semi-supervised graph node classification based on a simple nearest neighbor learning algorithm. In this experiment, the logarithmic forest distance performed best, while the RSP dissimilarity in many cases gave surprisingly poor results. Thus, it seems that the RSP dissimilarity can capture the global cluster structure of a graph quite well, but when focusing on individual, local distances, it is not so reliable. The logarithmic forest distance works well in the nearest neighbor learning task, but fails in many cases in the clustering task. The FE distance, however, performs well in both tasks. The FE distance is also defined in this paper on a solid foundation based on statistical thermodynamics, and it satisfies many desirable properties of a distance. One future plan is to use the different distance families in other network analysis tasks in order to characterize their differences more and give more insight on which distance is appropriate in which context. Also, we plan to extend even further the RSP framework for defining different graph node betweenness and robustness measures as well as temporal versions of the distances.

\appendix
\section{The CC distance is proportional to the CT distance}
\label{app:CT-CC}
For deriving this result, we refer to earlier literature. First, we call to mind a well-known result~\cite{Chandra-1989} that the commute time distance can be computed in terms of the pseudo-inverse of the graph Laplacian as
\begin{equation}
\label{eq:CTLaplacian}
\dist_{st}^{\mathrm{CT}} = (l_{ss}^{+} + l_{tt}^{+} - 2l_{st}^{+})\sum_{i,j = 1}^{n}a_{ij}.
\end{equation}
In addition, the authors in~\cite{FoussKDE-2005} derive a formula for computing the \emph{average first passage cost}, $o_{st}$, i.e.\ the expected cost of random walks from a node another. The formula (see~\cite{FoussKDE-2005}, Appendix B, Equation (18)) is given as
$o_{st} = \sum_{i=1}^{n}(l_{si}^{+} - l_{st}^{+} - l_{ti}^{+} + l_{tt}^{+})\sum_{j = 1}^{n}a_{ij}c_{ij}$.
From this we can obtain the commute cost distance $\dist_{st}^{\mathrm{CC}}$ by symmetrization:
\begin{align*}
\dist_{st}^{\mathrm{CC}}
&= o_{st} + o_{ts}
= \sum_{i = 1}^{n} (l_{si}^{+} - l_{st}^{+} - l_{ti}^{+} + l_{tt}^{+} + l_{ti}^{+} - l_{ts}^{+} - l_{si}^{+} + l_{ss}^{+})\sum_{j = 1}^{n}a_{ij}c_{ij} \\
&= (l_{ss}^{+} + l_{tt}^{+} - 2l_{st}^{+})\sum_{i,j = 1}^{n}a_{ij}c_{ij},
\end{align*}
which holds because the graph is assumed undirected. Comparing this result with Equation~(\ref{eq:CTLaplacian}) we see that the distances only differ from each other by a multiplying factor. Moreover, we see that this factor is
\[
\dfrac{\dist_{st}^{\mathrm{CC}}}{\dist_{st}^{\mathrm{CT}}} 
= \dfrac{\sum_{i,j=1}^{n}a_{ij}c_{ij}}{\sum_{i,j=1}^{n}a_{ij}} 
= \dfrac{\mathbf{e}^{\mathsf{T}} \! (\mathbf{A} \circ \mathbf{C}) \mathbf{e} }{\mathbf{e}^{\mathsf{T}} \! \mathbf{Ae}}.
\]





\bibliographystyle{elsarticle-num} 
\begin{footnotesize}
\bibliography{distfamilies}

\begin{thebibliography}{10}

\bibitem{Akamatsu-1996}
T.~Akamatsu.
\newblock Cyclic flows, markov process and stochastic traffic assignment.
\newblock {\em Transportation Research B}, 30(5):369--386, 1996.

\bibitem{pRes}
M.~Alamgir and U.~{von~Luxburg}.
\newblock Phase transition in the familiy of p-resistances.
\newblock In {\em Neural Information Processing Systems (NIPS)}, 2011.

\bibitem{Bavaud-2012}
F.~Bavaud and G.~Guex.
\newblock Interpolating between random walks and shortest paths: a path
  functional approach.
\newblock In K.~A. et~al., editor, {\em SocInfo 2012}, volume 7710 of {\em
  Lecture Notes in Computer Science}, pages 68--81. Springer, 2012.

\bibitem{ModernMDS}
I.~Borg and P.~Groenen.
\newblock {\em Modern multidimensional scaling: Theory and applications}.
\newblock Springer, 1997.

\bibitem{Brand-05}
M.~Brand.
\newblock A random walks perspective on maximizing satisfaction and profit.
\newblock {\em Proceedings of the 2005 SIAM International Conference on Data
  Mining}, 2005.

\bibitem{Chandra-1989}
A.~K. Chandra, P.~Raghavan, W.~L. Ruzzo, R.~Smolensky, and P.~Tiwari.
\newblock The electrical resistance of a graph captures its commute and cover
  times.
\newblock {\em Annual ACM Symposium on Theory of Computing}, pages 574--586,
  1989.

\bibitem{Chebotarev-2011}
P.~Chebotarev.
\newblock A class of graph-geodetic distances generalizing the shortest-path
  and the resistance distances.
\newblock {\em Discrete Applied Mathematics}, 159(5):295--302, 2011.

\bibitem{Chebotarev-2012}
P.~Chebotarev.
\newblock The walk distances in graphs.
\newblock {\em Discrete Applied Mathematics}, 160(10-11):1484--1500, 2012.

\bibitem{Chebotarev-1997}
P.~Chebotarev and E.~Shamis.
\newblock The matrix-forest theorem and measuring relations in small social
  groups.
\newblock {\em Automation and Remote Control}, 58(9):1505--1514, 1997.

\bibitem{Chebotarev-2002}
P.~Chebotarev and E.~Shamis.
\newblock The forest metric for graph vertices.
\newblock {\em Electronic Notes in Discrete Mathematics}, 11:98--107, 2002.

\bibitem{chung06}
F.~Chung and L.~Lu.
\newblock {\em Complex Graphs and Networks}.
\newblock American Mathematical Society, 2006.

\bibitem{Delvenne-2011}
J.-C. Delvenne and A.-S. Libert.
\newblock Centrality measures and thermodynamic formalism for complex networks.
\newblock {\em Physical Review E}, 83:046117, 2011.

\bibitem{Snell-1984}
P.~G. Doyle and J.~L. Snell.
\newblock {\em Random Walks and Electric Networks}.
\newblock The Mathematical Association of America, 1984.

\bibitem{Elkan2003}
C.~Elkan.
\newblock Using the triangle inequality to accelerate k-means.
\newblock In {\em ICML}, pages 147--153, 2003.

\bibitem{Estrada-2012}
E.~Estrada.
\newblock {\em The structure of complex networks: theory and applications}.
\newblock Oxford University Press, 2012.

\bibitem{Fortunato-2010}
S.~Fortunato.
\newblock Community detection in graphs.
\newblock {\em Physics Reports}, 486(3-5):75 -- 174, 2010.

\bibitem{FoussKDE-2005}
F.~Fouss, A.~Pirotte, J.-M. Renders, and M.~Saerens.
\newblock Random-walk computation of similarities between nodes of a graph,
  with application to collaborative recommendation.
\newblock {\em IEEE Transactions on Knowledge and Data Engineering},
  19(3):355--369, 2007.

\bibitem{BoP}
K.~Fran\c{c}oisse, I.~Kivim\"aki, A.~Mantrach, F.~Rossi, and M.~Saerens.
\newblock A bag of paths framework for community detection and semi-supervised
  classification.
\newblock Technical report, Universit{\'e} Catholique de Louvain, 2012.

\bibitem{Garcia-Diez-2011}
S.~Garcia-Diez, E.~Vandenbussche, and M.~Saerens.
\newblock A continuous-state version of discrete randomized shortest-paths,
  with application to path planning.
\newblock In {\em Decision and Control and European Control Conference
  (CDC-ECC), 2011 50th IEEE Conference on}, pages 6570 --6577, December 2011.

\bibitem{Gobel-1974}
F.~Gobel and A.~A. Jagers.
\newblock Random walks on graphs.
\newblock {\em Stochastic Processes and their Applications}, 2:311--336, 1974.

\bibitem{Klein-1993}
D.~J. Klein and M.~Randic.
\newblock Resistance distance.
\newblock {\em Journal of Mathematical Chemistry}, 12(1):81--95, 1993.

\bibitem{kolaczyk09}
E.~D. Kolaczyk.
\newblock {\em Statistical Analysis of Network Data}.
\newblock Springer, 2009.

\bibitem{Lewis09}
T.~G. Lewis.
\newblock {\em Network Science : Theory and Applications}.
\newblock Wiley, 2009.

\bibitem{ShortestToAllPath}
Y.~Li, Z.-L. Zhang, and D.~Boley.
\newblock The routing continuum from shortest-path to all-path: A unifying
  theory.
\newblock In {\em Proceedings of the 2011 31st International Conference on
  Distributed Computing Systems}, ICDCS '11, pages 847--856, Washington, DC,
  USA, 2011. IEEE Computer Society.

\bibitem{Newman-2006}
M.~Newman.
\newblock Modularity and community structure in networks.
\newblock {\em Proceedings of the National Academy of Sciences (USA)},
  103:8577--8582, 2006.

\bibitem{Newman2002}
M.~Newman and M.~Girvan.
\newblock Community structure in social and biological networks.
\newblock {\em Proceedings of the National Academy Science}, pages 7821--7826,
  2002.

\bibitem{newman10}
M.~E. Newman.
\newblock {\em Networks : An Introduction}.
\newblock Oxford University Press, 2010.

\bibitem{Page-1998}
L.~Page, S.~Brin, R.~Motwani, and T.~Winograd.
\newblock The pagerank citation ranking: Bringing order to the web.
\newblock {\em Technical Report 1999-0120, Computer Science Department,
  Stanford University}, 1999.

\bibitem{Peliti-2011}
L.~Peliti.
\newblock {\em Statistical Mechanics in a Nutshell}.
\newblock In a Nutshell. Princeton University Press, 2011.

\bibitem{CopelandsMethod}
D.~G. Saari.
\newblock Explaining all three-alternative voting outcomes.
\newblock {\em Journal of Economic Theory}, 87(2):313--355, 1999.

\bibitem{Saerens-2008}
M.~Saerens, Y.~Achbany, F.~Fouss, and L.~Yen.
\newblock Randomized shortest-path problems: Two related models.
\newblock {\em Neural Computation}, 21(8):2363--2404, 2009.

\bibitem{Shawe-Taylor-2004}
J.~Shawe-Taylor and N.~Cristianini.
\newblock {\em Kernel methods for pattern analysis}.
\newblock Cambridge University Press, 2004.

\bibitem{Taylor-1998}
H.~M. Taylor and S.~Karlin.
\newblock {\em An Introduction to Stochastic Modeling}.
\newblock Academic Press, 3rd edition, 1998.

\bibitem{Isomap}
J.~B. Tenenbaum, V.~d. Silva, and J.~C. Langford.
\newblock {A Global Geometric Framework for Nonlinear Dimensionality
  Reduction}.
\newblock {\em Science}, 290(5500):2319--2323, 2000.

\bibitem{Thelwall04}
M.~Thelwall.
\newblock {\em Link Analysis: An Information Science Approach}.
\newblock Elsevier, 2004.

\bibitem{vonLuxburg-2010}
U.~von Luxburg, A.~Radl, and M.~Hein.
\newblock Getting lost in space: large sample analysis of the commute distance.
\newblock {\em Proceedings of the 23th Neural Information Processing Systems
  conference (NIPS 2010)}, pages 2622--2630, 2010.

\bibitem{Wasserman-1994}
S.~Wasserman and K.~Faust.
\newblock {\em Social Network Analysis: Methods and Applications}.
\newblock Cambridge University Press, 1994.

\bibitem{SigCT}
L.~Yen, F.~Fouss, C.~Decaestecker, P.~Francq, and M.~Saerens.
\newblock Graph nodes clustering with the sigmoid commute-time kernel: A
  comparative study.
\newblock {\em Data and Knowledge Engineering}, 68(3):338 -- 361, 2009.

\bibitem{RSP}
L.~Yen, A.~Mantrach, M.~Shimbo, and M.~Saerens.
\newblock A family of dissimilarity measures between nodes generalizing both
  the shortest-path and the commute-time distances.
\newblock In {\em Proceedings of the 14th SIGKDD International Conference on
  Knowledge Discovery and Data Mining (KDD 2008)}, pages 785--793, 2008.

\bibitem{Zachary1977}
W.~W. Zachary.
\newblock An information flow model for conflict and fission in small groups.
\newblock {\em Journal of Anthropological Research}, (33):452--473, 1977.

\end{thebibliography}


\begin{thebibliography}{10}
\expandafter\ifx\csname url\endcsname\relax
  \def\url#1{\texttt{#1}}\fi
\expandafter\ifx\csname urlprefix\endcsname\relax\def\urlprefix{URL }\fi
\expandafter\ifx\csname href\endcsname\relax
  \def\href#1#2{#2} \def\path#1{#1}\fi

\bibitem{Wasserman-1994}
S.~Wasserman, K.~Faust, Social Network Analysis: Methods and Applications,
  Cambridge University Press, 1994.

\bibitem{Thelwall04}
M.~Thelwall, Link Analysis: An Information Science Approach, Elsevier, 2004.

\bibitem{chung06}
F.~Chung, L.~Lu, Complex Graphs and Networks, American Mathematical Society,
  2006.

\bibitem{Liben-Nowell-2007}
D.~Liben-Nowell, J.~Kleinberg, The link-prediction problem for social networks,
  Journal of the American Society for Information Science and Technology 58~(7)
  (2007) 1019--1031.

\bibitem{kolaczyk09}
E.~D. Kolaczyk, Statistical Analysis of Network Data, Springer, 2009.

\bibitem{Lewis09}
T.~G. Lewis, Network Science : Theory and Applications, Wiley, 2009.

\bibitem{newman10}
M.~E. Newman, Networks : An Introduction, Oxford University Press, 2010.

\bibitem{Lu-2011}
L.~L{\"u}, T.~Zhou, Link prediction in complex networks: A survey, Physica A:
  Statistical Mechanics and its Applications 390~(6) (2011) 1150 -- 1170.

\bibitem{Estrada-2012}
E.~Estrada, The structure of complex networks: theory and applications, Oxford
  University Press, 2012.

\bibitem{Gobel-1974}
F.~Gobel, A.~A. Jagers, Random walks on graphs, Stochastic Processes and their
  Applications 2 (1974) 311--336.

\bibitem{Klein-1993}
D.~J. Klein, M.~Randic, Resistance distance, Journal of Mathematical Chemistry
  12~(1) (1993) 81--95.

\bibitem{Snell-1984}
P.~G. Doyle, J.~L. Snell, Random Walks and Electric Networks, The Mathematical
  Association of America, 1984.

\bibitem{Chandra-1989}
A.~K. Chandra, P.~Raghavan, W.~L. Ruzzo, R.~Smolensky, P.~Tiwari, The
  electrical resistance of a graph captures its commute and cover times, Annual
  ACM Symposium on Theory of Computing (1989) 574--586.

\bibitem{RSP}
L.~Yen, A.~Mantrach, M.~Shimbo, M.~Saerens, A family of dissimilarity measures
  between nodes generalizing both the shortest-path and the commute-time
  distances, in: Proceedings of the 14th SIGKDD International Conference on
  Knowledge Discovery and Data Mining (KDD 2008), 2008, pp. 785--793.

\bibitem{Saerens-2008}
M.~Saerens, Y.~Achbany, F.~Fouss, L.~Yen, Randomized shortest-path problems:
  Two related models, Neural Computation 21~(8) (2009) 2363--2404.

\bibitem{BoP}
K.~Fran{\c{c}}oisse, I.~Kivim{\"a}ki, A.~Mantrach, F.~Rossi, M.~Saerens, A
  bag-of-paths framework for network data analysis, arXiv preprint
  arXiv:1302.6766.

\bibitem{Chebotarev-2011}
P.~Chebotarev, A class of graph-geodetic distances generalizing the
  shortest-path and the resistance distances, Discrete Applied Mathematics
  159~(5) (2011) 295--302.

\bibitem{Isomap}
J.~B. Tenenbaum, V.~d. Silva, J.~C. Langford, {A Global Geometric Framework for
  Nonlinear Dimensionality Reduction}, Science 290~(5500) (2000) 2319--2323.

\bibitem{Taylor-1998}
H.~M. Taylor, S.~Karlin, An Introduction to Stochastic Modeling, 3rd Edition,
  Academic Press, 1998.

\bibitem{Brand-05}
M.~Brand, A random walks perspective on maximizing satisfaction and profit,
  Proceedings of the 2005 SIAM International Conference on Data Mining.

\bibitem{radl2009resistance}
A.~Radl, U.~von Luxburg, M.~Hein, The resistance distance is meaningless for
  large random geometric graphs, in: Proc. Workshop on Analyzing Networks and
  Learning with Graphs, 2009.

\bibitem{vonLuxburg-2010}
U.~von Luxburg, A.~Radl, M.~Hein, Getting lost in space: large sample analysis
  of the commute distance, Proceedings of the 23th Neural Information
  Processing Systems conference (NIPS 2010) (2010) 2622--2630.

\bibitem{Akamatsu-1996}
T.~Akamatsu, Cyclic flows, markov process and stochastic traffic assignment,
  Transportation Research B 30~(5) (1996) 369--386.

\bibitem{Garcia-Diez-2011}
S.~Garcia-Diez, E.~Vandenbussche, M.~Saerens, A continuous-state version of
  discrete randomized shortest-paths, with application to path planning, in:
  Decision and Control and European Control Conference (CDC-ECC), 2011 50th
  IEEE Conference on, 2011, pp. 6570 --6577.

\bibitem{Elkan2003}
C.~Elkan, Using the triangle inequality to accelerate k-means, in: T.~Fawcett,
  N.~Mishra (Eds.), Machine Learning, Proceedings of the Twentieth
  International Conference (ICML 2003), August 21-24, 2003, Washington, DC,
  USA, AAAI Press, 2003, pp. 147--153.

\bibitem{Peliti-2011}
L.~Peliti, Statistical Mechanics in a Nutshell, In a Nutshell, Princeton
  University Press, 2011.

\bibitem{Fouss-2012}
F.~Fouss, K.~Fran\c{c}oisse, L.~Yen, A.~Pirotte, M.~Saerens, An experimental
  investigation of kernels on graphs for collaborative recommendation and
  semisupervised classification, Neural Networks 31 (2012) 53--72.

\bibitem{Bavaud-2012}
F.~Bavaud, G.~Guex, Interpolating between random walks and shortest paths: a
  path functional approach, in: K.~A. et~al. (Ed.), SocInfo 2012, Vol. 7710 of
  Lecture Notes in Computer Science, Springer, 2012, pp. 68--81.

\bibitem{Delvenne-2011}
J.-C. Delvenne, A.-S. Libert, Centrality measures and thermodynamic formalism
  for complex networks, Physical Review E 83 (2011) 046117.

\bibitem{Page-1998}
L.~Page, S.~Brin, R.~Motwani, T.~Winograd, The pagerank citation ranking:
  Bringing order to the web, Technical Report 1999-0120, Computer Science
  Department, Stanford University.

\bibitem{Klein-1998}
D.~Klein, H.-Y. Zhu, Distances and volumina for graphs, Journal of mathematical
  chemistry 23~(1) (1998) 179--195.

\bibitem{Chebotarev-1997}
P.~Chebotarev, E.~Shamis, The matrix-forest theorem and measuring relations in
  small social groups, Automation and Remote Control 58~(9) (1997) 1505--1514.

\bibitem{Chebotarev-2002}
P.~Chebotarev, E.~Shamis, The forest metrics for graph vertices, Electronic
  Notes in Discrete Mathematics 11 (2002) 98--107.

\bibitem{Chebotarev-2012}
P.~Chebotarev, The walk distances in graphs, Discrete Applied Mathematics
  160~(10-11) (2012) 1484--1500.

\bibitem{pRes}
M.~Alamgir, U.~{von~Luxburg}, Phase transition in the familiy of p-resistances,
  in: Neural Information Processing Systems (NIPS), 2011.

\bibitem{ModernMDS}
I.~Borg, P.~Groenen, Modern multidimensional scaling: Theory and applications,
  Springer, 1997.

\bibitem{Herbster-2010}
M.~Herbster, A triangle inequality for p-resistance, in: NIPS Workshop 2010:
  Networks Across Disciplines: Theory and Applications, 2010.

\bibitem{ShortestToAllPath}
Y.~Li, Z.-L. Zhang, D.~Boley, The routing continuum from shortest-path to
  all-path: A unifying theory, in: Proceedings of the 2011 31st International
  Conference on Distributed Computing Systems, ICDCS '11, IEEE Computer
  Society, Washington, DC, USA, 2011, pp. 847--856.

\bibitem{Estrada-2012-linalg}
E.~Estrada, The communicability distance in graphs, Linear Algebra and its
  Applications 436~(11) (2012) 4317--4328.

\bibitem{Estrada-2012-physrev}
E.~Estrada, Complex networks in the euclidean space of communicability
  distances, Physical Review E 85~(6) (2012) 066122.

\bibitem{bavaud2010euclidean}
F.~Bavaud, Euclidean distances, soft and spectral clustering on weighted
  graphs, in: Machine Learning and Knowledge Discovery in Databases, Springer,
  2010, pp. 103--118.

\bibitem{Lancichinetti-2009}
A.~Lancichinetti, S.~Fortunato, Benchmarks for testing community detection
  algorithms on directed and weighted graphs with overlapping communities,
  Phys. Rev. E 80 (2009) 016118.

\bibitem{SigCT}
L.~Yen, F.~Fouss, C.~Decaestecker, P.~Francq, M.~Saerens, Graph nodes
  clustering with the sigmoid commute-time kernel: A comparative study, Data
  and Knowledge Engineering 68~(3) (2009) 338 -- 361.

\bibitem{Shawe-Taylor-2004}
J.~Shawe-Taylor, N.~Cristianini, Kernel methods for pattern analysis, Cambridge
  University Press, 2004.

\bibitem{schoenberg1938metric}
I.~J. Schoenberg, Metric spaces and positive definite functions, Transactions
  of the American Mathematical Society 44~(3) (1938) 522--536.

\bibitem{gower1986metric}
J.~C. Gower, P.~Legendre, Metric and euclidean properties of dissimilarity
  coefficients, Journal of classification 3~(1) (1986) 5--48.

\bibitem{MarcoPCA}
M.~Saerens, F.~Fouss, L.~Yen, P.~Dupont, The principal components analysis of a
  graph, and its relationships to spectral clustering, Proceedings of the 15th
  European Conference on Machine Learning (ECML 2004). Lecture Notes in
  Artificial Intelligence, vol. 3201, Springer-Verlag, Berlin (2004) 371--383.

\bibitem{Strehl02clusterensembles}
A.~Strehl, J.~Ghosh, C.~Cardie, Cluster ensembles - a knowledge reuse framework
  for combining multiple partitions, Journal of Machine Learning Research 3
  (2002) 583--617.

\bibitem{Zhu-2009}
X.~Zhu, A.~Goldberg, Introduction to Semi-Supervised Learning, Synthesis
  Lectures on Artificial Intelligence and Machine Learning Series, Morgan \&
  Claypool, 2009.

\bibitem{Macskassy-2007}
S.~A. Macskassy, F.~Provost, Classification in networked data: A toolkit and a
  univariate case study, J. Mach. Learn. Res. 8 (2007) 935--983.

\bibitem{CopelandsMethod}
D.~G. Saari, Explaining all three-alternative voting outcomes, Journal of
  Economic Theory 87~(2) (1999) 313--355.

\bibitem{FoussKDE-2005}
F.~Fouss, A.~Pirotte, J.-M. Renders, M.~Saerens, Random-walk computation of
  similarities between nodes of a graph, with application to collaborative
  recommendation, IEEE Transactions on Knowledge and Data Engineering 19~(3)
  (2007) 355--369.

\end{thebibliography}
\end{footnotesize}

%
%
%
\end{document}